\documentclass[11pt]{article}
\usepackage[margin=1in]{geometry}
\usepackage[round]{natbib}
\usepackage{amsmath,amssymb,amsthm,mathtools}
\usepackage{graphicx,booktabs,xcolor,longtable,adjustbox,float}
\usepackage[hidelinks]{hyperref}
\hypersetup{pdftitle={Bounded-Order Non-Identifiability in Latent Structure Block Models: The Fiber in Closed Form},pdfauthor={Sie Hendrata Dharmawan and Peter Chin}}
\usepackage[capitalize,nameinlink]{cleveref}

\newtheorem{theorem}{Theorem}[section]
\newtheorem{proposition}[theorem]{Proposition}
\newtheorem{lemma}[theorem]{Lemma}
\newtheorem{corollary}[theorem]{Corollary}
\theoremstyle{definition}
\newtheorem{definition}[theorem]{Definition}
\newtheorem{example}[theorem]{Example}
\theoremstyle{remark}
\newtheorem*{remark}{Remark}

\newcommand{\E}{\mathbb{E}}
\newcommand{\arc}{s/\mathrm{arc}}
\newcommand{\status}[1]{}  %

\begin{document}

\title{Bounded-Order Non-Identifiability in Latent Structure Block Models:\\ The Fiber in Closed Form}
\author{Sie Hendrata Dharmawan \qquad Peter Chin\\[4pt]
Dartmouth College}
\date{}
\maketitle

\begin{abstract}
A graphon is determined by all of its homomorphism densities, but motif-based
methods use finitely many. We compute exactly what that truncation costs in a
natural family: latent structure block models whose positions lie on $K$
rays, equivalently degree-corrected block models of rank two. For a given
radial law, the configurations that agree on every graph with fewer than
$2K-1$ vertices form a one-dimensional fiber, which we give in closed form
with explicit endpoints; the friendship graph $F_{K-1}$ is its unique
edge-minimal separator, so no fixed motif order serves every $K$. The
separating signal is suppressed by a factor of at most $m_1^{2(K-1)}$, where
$m_1$ is the mean squared slope, about
$10^{-4}$ at $K=4$: detecting it from motif counts would take about
$10^{15}$ nodes in a single graph, at any edge density. The information is
nonetheless in the graph; the fiber belongs to the statistic, not the model.
Common spectral pipelines, which count no motifs, still slide along the fiber
by more than a third of its length at ordinary densities, with confidence
regions that almost never contain the truth, while model-based fits of the
same embedding do not. We give practitioners four checks: whether a fit is
at risk, what to run instead, why a confidence interval will not warn, and
how wide the remaining ambiguity is.
\end{abstract}

\section{Introduction}\label{sec:intro}
A graphon is determined by the homomorphism densities of all finite graphs
\citep{lovasz2006limits,borgs2008convergent,borgs2010moments,diaconis2008graph}.
This is the warrant behind every method that summarizes a network by motif
counts. In practice such methods count finitely many patterns, usually on a
handful of vertices, and the theory that licenses the truncation says little
about what a particular truncation costs. We compute the cost exactly in one
natural family, and then follow it through three ways of reading the same
graph: a bounded-order motif census, the full law of the graph, and the
spectral pipelines used in practice.

The family is the latent structure block model (LSBM), a random dot product
graph (RDPG) whose latent positions lie on $K$ rays through the origin
\citep{young2007random,athreya2018statistical,athreya2021estimation,sannapassino2022latent}.
A node's community fixes the slope of its ray and a random radius fixes its
position along it. Equivalently, it is a degree-corrected stochastic
blockmodel (DCSBM) \citep{karrer2011stochastic} whose degree parameters are
the radii and whose connectivity matrix $B_{kl}=1+s_ks_l$ has rank two; under
the DCSBM, communities appear in the spectral embedding as rays
\citep{sannapassino2022spectral}. We work in the symmetric random-slope
family: slopes come in reflection pairs with equal weights, and all
communities share one radial law. The symmetry is deliberate. It is the least
favorable setting for exhibiting a fiber, since it leaves the fewest free
directions; relaxing it does not remove the fiber, only its closed form
(\cref{sec:scope}). It also makes every homomorphism density, for a fixed radial law, a
polynomial in the moments $m_j$ of the squared slope. Identifiability under
motif counting becomes a question about a polynomial map, and the set of
configurations the census of order $m$ conflates with a given one is the
\emph{fiber} of the census map through it.

Our theorem (\cref{sec:model}) answers the question for every $K\ge3$. The
census of order $2K-2$ determines $m_1,\dots,m_{K-2}$ and nothing more, so
the fiber through an admissible configuration is a curve; we give it in
closed form as an arc whose endpoints are classical quadrature rules, show
that nothing else in the family matches, and identify the friendship graph
$F_{K-1}$ on $2K-1$ vertices as the unique edge-minimal separator. What is
blind is governed by degree rather than size: every graph of maximum degree
at most $2K-3$ has constant density along the fiber, however many vertices
it has. Because the threshold grows with $K$, no fixed motif order serves the
whole family. The moment-problem machinery is classical; what is new is the
reduction of motif matching to it, and the exhaustive, closed-form
description that results.

At the threshold the separation is present but minute, for a structural
reason. Any motif reads $m_j$ through terms at most $m_1^{2j}$, and
$F_{K-1}$ attains the bound; at $K=4$ that is about $10^{-4}$, and two
configurations differing by $45.1\%$ in $m_3$ differ by $7.6\times10^{-7}$
in $t(F_3)$. Detection from motif counts would take about $2\times10^{15}$
nodes in a single graph ($10^{12}$--$10^{13}$ even pooling $200$ graphs), and
the obstruction is invariant to edge density (\cref{sec:census}). Lower-order statistics are blinder still: trees, and
hence message-passing graph neural networks, which distinguish graphons
exactly as tree densities do \citep{boker2023fine}, see nothing of the slopes
at all (\cref{cor:mpnn}). Yet the information is in the graph: its law
identifies the configuration, and model-based fits recover the position along
the fiber (\cref{sec:oracle}). The fiber belongs to the pair (model, statistic), not
to the model.

The fiber resurfaces where nothing counts motifs. Spectral pipelines for the
DCSBM embed the adjacency matrix, reduce each node to a direction (a ratio of
eigenvector entries, an angle, or a unit vector), cluster, and read the
parameters off the clusters
\citep{jin2015fast,qin2013regularized,lei2015consistency,sannapassino2022spectral}.
Adjacency spectral embedding is consistent; the loss is in the summary taken
afterward. On graphs simulated from the family, in an intermediate
regime the error of such pipelines collapses onto the fiber and slides along
it by more than a third of its length, with replicate spread $5$--$14$ times
smaller than the displacement, and nominal $95\%$ regions contain the truth in
no cell of $17$ of $18$ groups (\cref{sec:spectral}). Concretely, a
community holding $3.75\%$ of the nodes is reported at about $17\%$, every
replicate lies within $2\%$ of the arc of the truth's census fiber, and no
nominal $95\%$ region contains the truth. Two of three published pipelines
slide this way, by up to $58\%$ of the arc; the third avoids the slide in that
regime but fails bimodally near the fiber's endpoint, and a standard
preprocessing filter restores the slide. Model-based fits, given the ray
family and the embedding's noise model---a ray mixture under the ASE covariance, and
the LSBM authors' Bayesian sampler---are accurate at every cell tested.

We present this as a caution with a remedy: the algorithms are correct, the
interpretation of their output is at risk, and the risk can be checked and
avoided. \Cref{sec:practitioner} gives four checks---whether a fit sits in
the displacement regime, what to run instead, why a confidence interval will
not raise the alarm, and how wide the remaining ambiguity is.

\paragraph{Contributions.}
(i) A non-identifiability theorem for every $K\ge3$ with the fiber in closed
form: existence, shape, size, and cost. (ii) A hierarchy of what bounded-order
statistics see: message passing nothing, graphs of maximum degree at most
$2K-3$ nothing along the fiber, $F_{K-1}$ the fiber at a $10^{-4}$
suppression. (iii) Evidence that no information barrier is responsible. (iv)
An account of spectral pipelines on the fiber: published methods are
displaced along it with silent confidence regions; a regime map; and
practitioner checks.

\section{Related Work}\label{sec:related}
\paragraph{RDPGs and spectral estimation.}
Adjacency spectral embedding (ASE) is consistent for RDPG latent positions up
to orthogonal transformation \citep{sussman2012consistent}, clusters
perfectly for large $n$ \citep{lyzinski2014perfect}, and obeys row-wise
central limit theorems \citep{athreya2016limit,tang2018limit} and
two-to-infinity bounds \citep{cape2019two}; see also
\citet{rubindelanchy2022statistical}. For latent structure, one embeds and
then estimates the support of the embedded cloud
\citep{athreya2021estimation,trosset2020learning,rubindelanchy2020manifold,sannapassino2022latent}.
For the DCSBM, spectral clustering reduces each node to a direction: SCORE
takes entrywise ratios of eigenvectors \citep{jin2015fast}, others normalize
rows \citep{qin2013regularized,lei2015consistency} or pass to spherical
coordinates \citep{sannapassino2022spectral}. Their guarantees concern
community labels; we study the parameters read off the clusters. These are
consistency and first-order statements. Plug-in summaries of noisy
embeddings carry second-order bias: noise inflates second moments, the
attenuation effect of errors-in-variables models
\citep{fuller1987measurement,carroll2006measurement}, and higher-order
spectral theory with explicit bias terms is emerging
\citep{xie2024higher,zhang2022edgeworth,levin2019bootstrapping}.
What is new here is the amplifier: a flat direction of the population
objective turns a second-order bias into an $O(1)$ displacement along the
fiber.

\paragraph{Graphons, densities, identifiability.}
Densities of all finite graphs determine a graphon up to weak isomorphism
\citep{lovasz2006limits,borgs2010moments}, but the inverse counting lemma
controls cut distance from densities up to order $k$ only at rate
$O(1/\sqrt{\log k})$ \citep{borgs2008convergent,lovasz2012large}.
Densities are a determinate but badly conditioned chart. For the $K$-block
SBM, \citet[Thm.~2]{bickel2011method} show that wheel densities identify the
parameters when $\pi,F\pi,\dots,F^{K-1}\pi$ are linearly independent, which
excludes equal expected degrees across blocks. Our family sits in the
analogue of the excluded regime: with $\E L=(\E x_1,0)$ and $\E[LL^\top]$
diagonal, the iterated degree function is
$T^j(1)(L)=x_1(\E x_1)(\E x_1^2)^{j-1}$, so every wheel density is a
functional of the radial law alone, constant across the whole slope family.

Elsewhere, RDPG latent laws are identifiable from the full graph law
\citep{lei2021network}, block models generically so
\citep{allman2011parameter}, and tensor methods recover block parameters
from low-order statistics under rank conditions \citep{anandkumar2014tensor}.
Message-passing neural networks are bounded by the 1-WL test
\citep{xu2019powerful,morris2019weisfeiler}, which on finite graphs is
equivalent to tree homomorphism counts \citep{dvorak2010recognizing,dell2018lovasz}
and on graphons to tree densities \citep{grebik2022fractional,boker2023fine}.
We use ``fiber'' as in algebraic statistics, where non-identifiability is a
positive-dimensional fiber of the parametrization \citep{drton2009lectures}.

\paragraph{Moments and motif counts.}
Our construction is classical: the truncated Stieltjes problem, Hankel and
Jacobi matrices, Gauss--Radau rules
\citep{shohat1943problem,akhiezer1965classical,gautschi2004orthogonal},
applied to mixtures since \citet{lindsay1989moment}. Hankel condition
numbers grow exponentially \citep{tyrtyshnikov1994how,beckermann2000condition},
and point-mass recovery from noisy moments is fragile
\citep{moitra2015super}; determinacy and conditioning come apart, which is
the census's failure. Motif counts are old network summaries
\citep{milo2002network,przulj2007biological,shervashidze2009efficient} with
mature distribution theory
\citep{bickel2011method,bhattacharyya2015subsampling,maugis2020testing}; we ask
what the densities themselves determine.

\section{Model and Main Theorem}\label{sec:model}\label{sec:generative}
\paragraph{Generative model.}
Fix $K\ge3$ and $C>0$. A configuration is
$\Theta=((w_1,\dots,w_K),(s_1,\dots,s_K),R)$ with weights $w_i>0$,
$w_i=w_{K+1-i}$, $\sum_iw_i=1$; slopes $s_1>\dots>s_K$ with
$s_i=-s_{K+1-i}$ (so $s_{k+1}=0$ when $K=2k+1$); and a radial law $R$ on
$(0,C)$ with moments $\rho_d:=\E_R[x^d]$. Each node $u$ draws a community
$c_u$ with $\Pr[c_u=i]=w_i$ and $x_{1,u}\sim R$ independently, and sets
$L_u=x_{1,u}(1,s_{c_u})$. Nodes $u,v$ are joined independently with
probability
\begin{equation}\label{eq:kernel}
W(u,v)=\langle L_u,L_v\rangle=x_{1,u}x_{1,v}\,(1+s_{c_u}s_{c_v}).
\end{equation}
This is an LSBM with $K$ rays through the origin, and a DCSBM with degree
parameters $x_{1,u}$ and rank-two connectivity $B_{kl}=1+s_ks_l$. Writing $t_i=s_i^2$,
$W\in[0,1]$ for all pairs iff $t_1\le T:=\min(1,C^{-2}-1)$, which needs $C<1$; we call such
configurations \emph{admissible}. By reflection symmetry the odd slope
moments vanish; $m_j:=\E[s^{2j}]$ are the moments of the squared slope.

\paragraph{The census factorizes.}
Let $t(F;\Theta)$ be the homomorphism density of $F$ under~\eqref{eq:kernel};
injective motif counts are unbiased for it (\cref{prop:unbiased}), and
induced (graphlet) densities carry the same information order by order. The
\emph{census of order $m$}, $\mathrm{Cen}_m(\Theta)$, collects $t(F;\Theta)$
over graphs with $|V(F)|\le m$. Call $S\subseteq E(F)$ \emph{even} if every
vertex has even degree in $(V(F),S)$, and write $\mathcal{E}(F)$ for these
sets.

\begin{lemma}[Factorization and even-subgraph expansion]\label{lem:reduction}
For every graph $F$,
\[
t(F;\Theta)=\prod_{v\in V(F)}\rho_{\deg_F v}\;
\sum_{S\in\mathcal{E}(F)}\;\prod_{v\in V(F)} m_{\deg_S(v)/2}.
\]
\end{lemma}
\noindent The radii factor out because $\prod_{uv}x_ux_v=\prod_vx_v^{\deg v}$;
expanding $\prod_{uv}(1+s_us_v)$ and discarding odd slope moments leaves the
even subgraphs, as in the high-temperature expansion of statistical mechanics
\citep{vanderwaerden1941lange} (\cref{lem:factorization,thm:even_expansion}). Forests are
therefore blind to the slopes: edge density, degrees, stars and paths depend on
$R$ alone. A cycle sees $1+m_1^{\ell}$. What a graph can see is recorded by its
\emph{slope index} $h(F):=\max_{S\in\mathcal E(F)}\max_v\deg_S(v)/2$, the
highest moment in its expansion (\cref{def:slope_index}); $h(F)\le\lfloor
\Delta(F)/2\rfloor$, and $h(F)\ge j$ forces $|V(F)|\ge2j+1$
(\cref{prop:slope_index_bounds}). Because every coefficient is nonnegative
there is no cancellation, which gives a clean transfer principle.

\begin{proposition}[Motif transfer]\label{prop:transfer_main}
Let $\Theta_1,\Theta_2$ share $R$. (i) If $m_d(\Theta_1)=m_d(\Theta_2)$ for
$d\le j$, then $t(F;\Theta_1)=t(F;\Theta_2)$ whenever $h(F)\le j$. (ii) If
they agree for $d<j$ and $m_j(\Theta_1)>m_j(\Theta_2)$, then
$t(F;\Theta_1)>t(F;\Theta_2)$ whenever $h(F)=j$.
\end{proposition}
\noindent This is \cref{thm:motif_transfer}. Motif matching is thus moment
matching, graded by $h$; the question becomes which slope laws share
$m_0,\dots,m_{K-2}$.

\begin{corollary}[Message passing sees no slope]\label{cor:mpnn}
Configurations sharing $R$ have equal densities for every tree. Hence they are
indistinguishable by the continuous 1-WL test on graphons, and by every
message-passing network without node features, with order-normalized sum
aggregation and mean readout, in the graphon limit \citep{boker2023fine};
under refinement, every node's color is a function of its radius alone.
\end{corollary}
\noindent A tree has no nonempty even edge set, so \cref{lem:reduction} gives
$t(T;\Theta)=\prod_v\rho_{\deg v}$. Directly: a node's degree is
$x\rho_1(1+s_c\E[s])=x\rho_1$, and each round of refinement averages the
neighbors' slopes against a mean-zero law (\cref{app:new}). Only $\E[s]=0$ and
a community-independent radius are used. Higher-order variants see more,
since $F_{K-1}$ has treewidth $2$ \citep{dell2018lovasz}, but inherit the
suppression of \cref{prop:bound}.

\paragraph{The fiber.}
Let $K=2k$. From the matched moments form the Hankel matrix
$\mathbb{M}_k=(m_{a+b-2})_{a,b\le k}$; its Cholesky factor $U$
($\mathbb{M}_k=U^\top U$) gives the Jacobi coefficients $a_0=m_1,a_1,\dots,a_{k-2}$ and $b_1,\dots,b_{k-1}>0$, i.e.\ the
three-term recurrence of the monic orthogonal polynomials $f_j$ of the
matched moments (\cref{def:construction_even,prop:a_b_equivalence}). For
$\xi\in\mathbb{R}$ let $J(\xi)$ be the $k\times k$ tridiagonal matrix with
diagonal $(a_0,\dots,a_{k-2},\xi)$ and off-diagonal $(b_1,\dots,b_{k-1})$,
with eigenvalues $t_1(\xi)>\dots>t_k(\xi)$ and unit eigenvectors $v_i$. The
configuration $\Theta_\xi$ has slopes $\pm\sqrt{t_i}$, weights
$w_i=\tfrac12(v_i)_1^2$ on each ray of a reflected pair, and radial law $R$.
For $K=2k+1$ the middle ray has slope $0$, and the same construction is
applied to the size-biased moments $m'_d=m_{d+1}/m_1$, giving $J'(\xi)$; the
outer rays get weights $m_1(v'_i)_1^2/(2t'_i)$ and the middle ray the
remainder $w_{k+1}=1-m_1\,e_1^\top J'(\xi)^{-1}e_1$
(\cref{def:construction_odd,lem:middle_weight}). Write
$\Phi(\tilde\Theta)=\{\Theta_\xi:\xi\in(\xi^-,\xi^+]\}$.

\begin{theorem}\label{thm:main}
Let $K\ge3$, fix $R$, and let $\tilde\Theta$ be admissible with $K$
communities.
\begin{enumerate}\itemsep1pt
\item[(a)] \emph{Fiber.} $\Phi(\tilde\Theta)$ is exactly the set of
admissible $K$-community configurations with radial law $R$ whose
$m_0,\dots,m_{K-2}$ agree with those of $\tilde\Theta$. The interval
$(\xi^-,\xi^+]$ is nondegenerate, and $m_{K-1}(\Theta_\xi)=\sigma_K\xi+c_K$
with $\sigma_K>0$, so distinct $\xi$ give distinct configurations.
\item[(b)] \emph{Endpoints.} For $K=2k$,
\[
\xi^-=-b_{k-1}^2\tfrac{f_{k-2}(0)}{f_{k-1}(0)},\qquad
\xi^+=T-b_{k-1}^2\tfrac{f_{k-2}(T)}{f_{k-1}(T)},
\]
the Gauss--Radau rules of the matched law with a node fixed at $0$ and at $T$
\citep{golub1973some}.
For $K=2k+1$, $\xi^+$ is the same expression in the primed coefficients,
and $\xi^-=\sum_{j=0}^{k-1}\alpha_j-\sum_{j=0}^{k-2}a'_j$, where $\alpha_j$
are the recurrence coefficients of the matched law: the $k$-point Gauss rule,
at which the middle ray loses its mass.
\item[(c)] \emph{Matched motifs.} $t(F;\cdot)$ is constant on
$\Phi(\tilde\Theta)$ for every $F$ with $h(F)\le K-2$: in particular every
graph on at most $2K-2$ vertices, and every graph of maximum degree at most
$2K-3$, however large.
\item[(d)] \emph{Breakers.} For every $F$ with $h(F)=K-1$, $t(F;\cdot)$ is strictly increasing
along the fiber. A graph on $2K-1$ vertices separates two fiber points iff it
has a vertex adjacent to all others whose neighbourhood induces components of
even order; among these, the friendship graph $F_{K-1}$ is the unique one with
fewest edges. Its density is affine in $\xi$ with slope
\[
\rho_{2K-2}\,\rho_2^{2K-2}\,m_1^{2K-2}\,\sigma_K .
\]
\end{enumerate}
\end{theorem}

\begin{corollary}[The census sees exactly the fiber]\label{cor:census_char}
An admissible $K$-community configuration with radial law $R$ agrees with
$\tilde\Theta$ on $\mathrm{Cen}_{2K-2}$ iff it lies in $\Phi(\tilde\Theta)$.
\end{corollary}
\noindent If $\Theta\notin\Phi(\tilde\Theta)$, some $m_j$ with $1\le j\le
K-2$ differs; at the first such $j$, the friendship graph $F_j$ (on
$2j+1\le2K-3$ vertices, $h=j$) separates by \cref{prop:transfer_main}(ii)
(\cref{app:new}).

\begin{corollary}[No fixed motif order]\label{cor:nofixed_main}
For every $m$ there are $K$ and a nondegenerate fiber on which
$\mathrm{Cen}_m$ is constant (\cref{cor:diagonal}).
\end{corollary}

\paragraph{The smallest case.} At $K=3$, $J'=[\xi]$: the outer rays have
squared slope $\xi$ and weight $m_1/(2\xi)$ each, the middle ray weight
$1-m_1/\xi$, and the fiber is $\xi\in(m_1,T]$. The census of order $4$ is
constant along it, while $m_2=m_1\xi$ and the density of the bowtie $F_2$
increase. At $\xi\downarrow m_1$ the middle ray empties and the configuration
collapses to two rays; at $\xi=T$ the outer rays reach the admissibility
boundary.

\paragraph{Why the corner is free.}
The Cholesky factor makes the Gram matrix $\mathbb{M}_k$ of
$1,x,\dots,x^{k-1}$ orthonormal, and $J$ is the shifted-by-one Hankel matrix in
that basis (\cref{prop:insight}). The data fill $\mathbb{M}_k$ completely and
its shift everywhere except the bottom-right corner, which is $\xi$. Any matching
configuration yields the same $J$ up to its corner (\cref{lem:f_even,cor:uniqueness}). Since
$\partial J/\partial\xi=e_ke_k^\top$, every eigenvalue increases with $\xi$
and $\sum_i\mathrm{d}t_i/\mathrm{d}\xi=1$ (\cref{prop:monotone_transport}); by
interlacing only $t_1$ can exceed $T$ and only $t_k$ can reach $0$, which
gives the endpoints (\cref{thm:xi_admissibility}). For odd $K$ the middle
weight decreases in $\xi$ and diverges to $-\infty$ as $t'_k\downarrow0$, so
it vanishes first, at the Gauss rule (\cref{thm:xi_admissibility_odd}). The
breaker statements are combinatorial: on $2K-1$ vertices, a vertex of
$S$-degree $2K-2$ must be adjacent to all others, and its neighbours need odd
degree inside the neighbourhood, which by the handshake lemma forces
even-order components; fewest edges means a perfect matching, i.e.\ $F_{K-1}$
(\cref{prop:minimal_order,thm:breaker}). The machinery is classical: Krein's
canonical representations \citep{krein1977markov}, quasi-orthogonal
polynomials \citep{shohat1937mechanical}, and Gauss--Radau rules computed by
moving the corner \citep{golub1969calculation,golub1973some}.
The contribution is the reduction of motif matching to this truncated moment
problem, and the exhaustive closed-form description it yields.

\section{Geometry of the Fiber at $K=4$}\label{sec:geometry}
At $K=4$ ($k=2$) everything is explicit. With $b_1^2=m_2-m_1^2$,
$J(\xi)=\bigl(\begin{smallmatrix}m_1&b_1\\ b_1&\xi\end{smallmatrix}\bigr)$,
the squared-slope law is $\alpha_1\delta_{\tau_1}+(1-\alpha_1)\delta_{\tau_2}$
with $\tau_1<\tau_2$ the eigenvalues, $\alpha_1=(\tau_2-m_1)/(\tau_2-\tau_1)$,
and $\xi^-=b_1^2/m_1$, $\xi^+=T-b_1^2/(T-m_1)$. We parametrize configurations
by $\theta=(\tau_1,\tau_2,\alpha_1)$ and report position along the fiber as
$\arc$, the fraction of arc length---never as $\xi$, which traverses the
curve at non-constant speed. Experiments use the uniform radial law on
$(0,C)$, for which the mean degree is $nC^2/4$.

Through the configurations of \cref{tab:pair} ($m_1=0.21744$,
$m_2=0.05863$) and at $C=0.75$ ($T=7/9$), the fiber has length $1.0571$ in
$\theta$; its open end has $\tau_1\to0$ ($\alpha_1\approx0.194$) and its
closed end $\tau_2=T$ ($\alpha_1\approx0.965$); its best-fit plane captures
$99.96\%$ of its variance. For $C\le1/\sqrt2$, $T=1$ and the admissible arc
extends to length $1.280$; we normalize by $1.0571$ at every density, and
configurations beyond it ($s/\mathrm{arc}>1$) are used only at $C\le1/\sqrt2$.
Since $\alpha_1$ rises along the arc, the minority weight
$w_{\min}=\min(\alpha_1,1-\alpha_1)$ is two-to-one in position. We name five
configurations: LO ($w_{\min}=0.211$), pair A ($0.255$) and MID ($0.314$) on the
ascending branch, and HI ($0.0375$) and pair B ($0.036$) on the descending
branch, where the minority sits on the steep ray.

\begin{table}[t]
\caption{Two configurations on one $K=4$ fiber. The census matches below
order $7$ exactly; the separator reads a $45\%$ moment gap as $10^{-6}$.}
\label{tab:pair}\centering\small
\begin{tabular}{lccc}
\toprule
& pair A & pair B & rel.\ diff.\\
\midrule
$m_1$ & $0.21744283$ & $0.21744283$ & $0$ (50 digits)\\
$m_2$ & $0.05862892$ & $0.05862892$ & $0$ (50 digits)\\
$m_3$ & $0.01632560$ & $0.02369236$ & $45.1\%$\\
$t(F_3)$ & --- & --- & $7.551\times10^{-7}$\\
\bottomrule
\end{tabular}
\end{table}

The slope laws of pairs A and B differ by $16\%$ of the support range in Wasserstein
distance ($W_1=0.1176$): matching low-order densities says little about
graphon proximity, as the logarithmic inverse counting lemma predicts.

\section{The Census Route}\label{sec:census}
The suppression in \cref{tab:pair} is structural, not a poor choice of motif.
\begin{proposition}\label{prop:bound}
In the expansion of \cref{lem:reduction}, every term containing $m_j$
($j\ge1$) is at most $m_j\,m_1^{2j}$, and $F_j$ attains the bound.
\end{proposition}
\noindent A vertex of $S$-degree $2j$ has $2j$ distinct neighbours, each of
even $S$-degree at least $2$, and $m_r\le m_1$ for $r\ge1$ because the
squared slopes lie in $[0,1]$; every other factor is at most $m_0=1$
(\cref{app:new}). The bound is per term; the number of terms carrying $m_j$
is a fixed combinatorial count of $F$.

So the census reads the first unmatched moment $m_{K-1}$ through
$m_1^{2(K-1)}\approx10^{-4}$ at $K=4$: the best-conditioned census access to
$m_{K-1}$ that exists. To measure what this costs without committing to an
estimator, we sample $200$ graphs from each of two fiber configurations at
$n\in\{2000,4000,8000,16000,32000\}$, estimate $t(F_{K-1})$ per graph from
homomorphism counts, and form a two-sample $z$-statistic with the empirical
across-replicate standard error; the degenerate-map bias of homomorphism
counts cancels exactly in this difference (\cref{app:census}). The standard
deviation of the per-graph estimate decays as $n^{-0.49}$, as latent sampling
predicts, and we extrapolate it to the $n^*$ at which $|\Delta|/\mathrm{SE}=2.8$
(power $0.8$ at two-sided $\alpha=0.05$; \cref{tab:census}). Across four
fitting windows $n^*$ at $K=4$ ranges over $2.7\times10^{12}$--$9.7\times10^{12}$,
so we quote an order. These figures pool $200$ graphs per configuration; with
one graph per configuration $n^*$ rises about $200$-fold, to roughly
$2\times10^{15}$ at $K=4$. The extrapolation assumes the $n^{-1/2}$ decay of a
nondegenerate $U$-statistic persists; the graphs stay dense as $n$ grows, so no
sparse-graph transition intervenes. There is no
direct detection at any $K$ or $n$. Without extrapolating, the perturbation
needed for a detectable effect is about $100\times$ larger through the census
than through the spectrum ($78\times$ at $n=2000$, $123\times$ at
$n=32{,}000$). Sparsity does not help: for a scale family of radial laws $\rho_d\propto
C^d$, so by \cref{lem:reduction} the relative separation is exactly invariant
to $C$, and the measured
floor-to-signal ratio is flat to within $1.4\times$ from $C=0.75$ to $0.05$.

\begin{table}[t]
\caption{Census route by $K$ (threshold order $2K-1$): moment and density
separation along the fiber, and the extrapolated $n^*$ with $200$ graphs or
one graph per configuration.}\label{tab:census}
\centering\small
\begin{tabular}{ccccc}
\toprule
& & & \multicolumn{2}{c}{$n^*$ (extrapolated)}\\
\cmidrule(lr){4-5}
$K$ & $\Delta m_{K-1}$ & $\Delta t(F_{K-1})$ & 200 graphs & 1 graph\\
\midrule
4 & $45.1\%$ & $7.6\times10^{-7}$ & $10^{12}$--$10^{13}$ & $2\times10^{15}$\\
5 & $3.81\%$ & $2.3\times10^{-8}$ & $2\times10^{15}$ & $3\times10^{17}$\\
6 & $1.74\%$ & $3.6\times10^{-9}$ & $3$--$6\times10^{16}$ & $4\times10^{18}$\\
\bottomrule
\end{tabular}
\end{table}

The moment inversion the theory suggests ($m_2,m_3$ from $t(F_2),t(F_3)$, then
Prony) produced no valid slope law in any of $12{,}600$ replicates: dividing by
$m_1^4$ and $m_1^6$ amplifies noise $447\times$ and $9{,}461\times$. A
constrained variant returns valid fits, all on the boundary of the admissible
set and displaced by up to $72\%$ of the arc (\cref{app:census}).

\section{No Information Barrier}\label{sec:oracle}
The census cannot separate points of the fiber; the graph can. At the
population level, the law of an RDPG determines the distribution of latent
positions up to an orthogonal transformation \citep{lei2021network}. Here
$\E L=(\rho_1,0)$ pins it to the reflection
$x_2\mapsto-x_2$, which maps the symmetric family to itself, so distinct fiber
points induce distinct graph laws.

At finite $n$, model-based fits recover the position along the fiber from
the observed graph alone: RayEM to within $1\%$ of the arc and the LSBM
authors' sampler to within $2.3\%$, at every cell tested
(\cref{sec:spectral}, \cref{tab:pipelines}). As a check that involves no
estimation, the held-out log-loss with the true latent positions snapped to each candidate's rays picks the
generating configuration in all $36$ cells ($K=4,5,6$, down to mean degree
$5$), and its margin matches the analytic population log-loss gap at ratios
$0.93$--$1.04$ (\cref{app:scores}). The obstruction belongs to the census,
not to the graph.

\section{The Spectral Pipeline}\label{sec:spectral}
\paragraph{Estimators.} \emph{P1-cluster} is the embed-then-cluster pipeline
in its simplest form: the two-dimensional ASE with an observable-only
canonicalization, a per-node slope $\hat T=\hat x_2/\hat x_1$ (in $d=2$,
SCORE's eigenvector ratio up to a constant), a small-radius filter, and
one-dimensional $2$-means on $|\hat T|$; atoms are cluster means of $\hat T^2$
and the weight is the cluster proportion. \emph{P1-Prony}, the classical
moment-matching alternative, applies Prony's method to the first three
moments of $\hat T^2$. Both are handed $K$, $d$ and the reflection symmetry.

\begin{figure*}[t]
\centering
\includegraphics[width=0.83\textwidth]{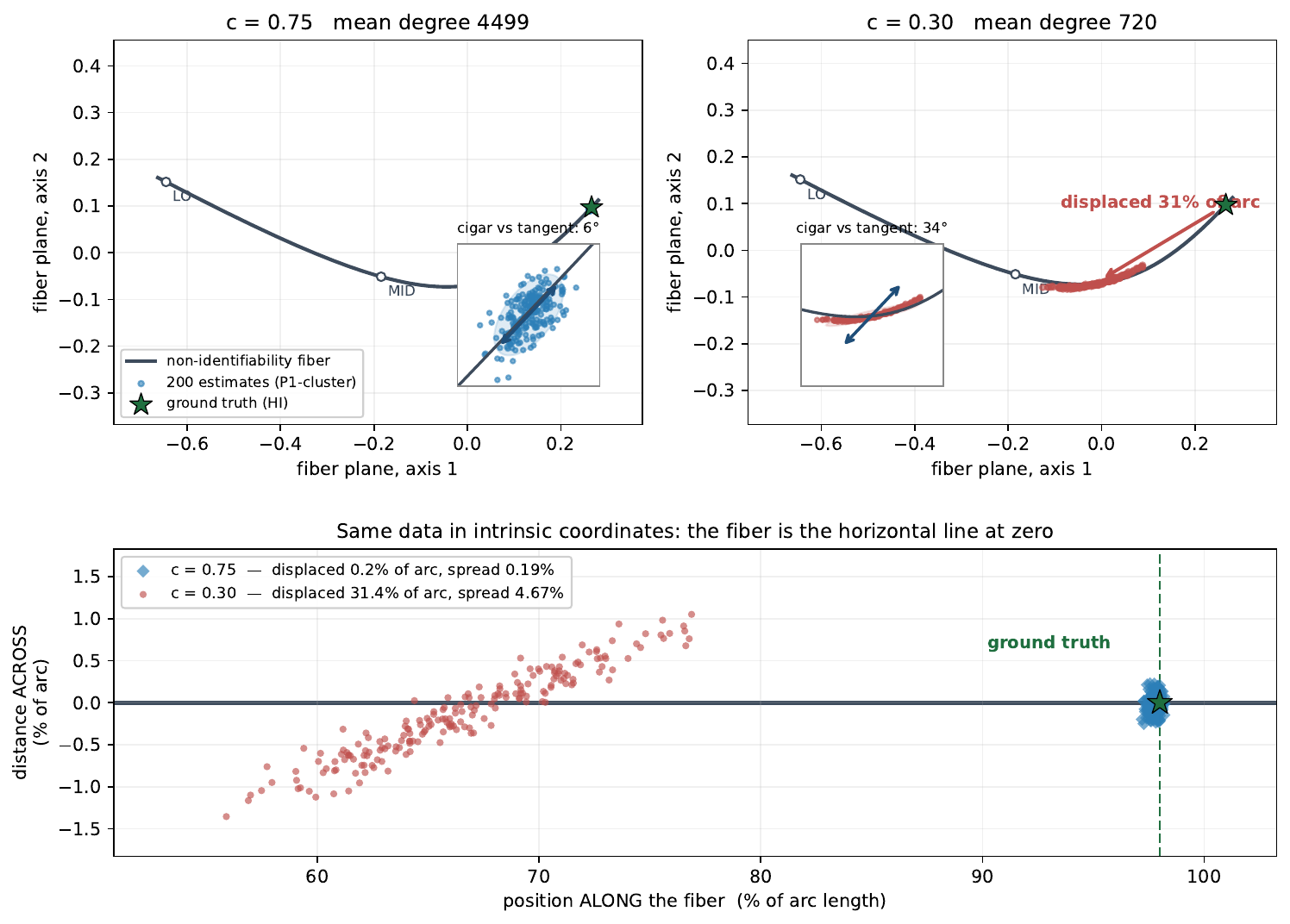}
\caption{P1-cluster at configuration HI ($w_{\min}=0.0375$), $n=32{,}000$.
Top: the fiber and estimates in the fiber's best-fit plane at two densities.
Bottom: the same estimates in intrinsic coordinates (position along the fiber
versus signed distance across it). At mean degree $720$ every estimate lies
close to the curve yet is displaced $31\%$ of the arc along it.}\label{fig:cigar}
\end{figure*}

\paragraph{Bias, not spread.} At $C=0.3$, $n=32{,}000$ (mean degree $720$)
P1-cluster lands $33\times$ closer to the fiber than to the truth; all $200$
replicates lie within $0.02$ of it in $\theta$ ($1.9\%$ of the arc) while displaced along
it (\cref{fig:cigar}). The displacement is bias, $7$--$15$ times the spread
at the named configurations (\cref{tab:bias}) and $5$--$14$ times across all
on-curve cells displaced by at least $5\%$ (\cref{app:tables}). The estimate is precise, on the curve, and
wrong. It is not an implementation error: the same code recovers $\alpha_1$ to
$0.5\%$ at $C=0.75$, and other initializations leave HI at $29.9$--$31.4\%$.

\paragraph{Why the census fiber.} At $K=4$ the census fiber is exactly the
level set of $(m_1,m_2)$ (\cref{thm:main}(a)), so ``on the fiber, displaced
along it'' could be read as ``$\hat m_1,\hat m_2$ accurate, $\hat m_3$ not''.
That reading is the point rather than an objection: the theorem says this
level set is all that low-order statistics determine, and any estimator that
fits the low moments well but the next one poorly inherits it. What the theorem adds is the set's geometry, a curve with known
endpoints and length, which makes the displacement measurable and bounded.
The error follows that curve, not a fixed ill-conditioned direction: the mean error tracks the local tangent to within $16$--$24^\circ$ at five
configurations whose tangents differ by up to $75^\circ$; moving the truth
off the fiber destroys the alignment ($15.7^\circ$ on it, $66.5^\circ$ at
offset $0.35$); and RayEM, fitted to the same embedding, is not displaced.

\paragraph{Two tilts, one amplifier.} A flat direction alone produces wide
unbiased scatter; displacement also needs a tilt, a systematic estimator
error, which the fiber amplifies because across it the objective climbs and
along it nothing resists. For P1-cluster the tilt is a partition error: the
fitted minority weight is inflated $1.11\times$ at balanced weights and
$4.4\times$ at $w_{\min}=0.0375$. For P1-Prony it is a moment error: ASE noise
inflates $\hat m_3$ by $40$--$64\%$, and the displacement is toward larger
$m_3$ in $229$ of $276$ cells. Both are second-order ASE bias
($\E\hat S=m_1+\sigma^2$, $\sigma^2=O(n^{-1})$; \cref{app:spectral}).

\paragraph{Published pipelines.} We ran three published DCSBM pipelines on the
same embeddings, reading parameters off their clusters with one shared
plug-in (\cref{app:pipelines}): row-normalized spectral clustering
\citep{qin2013regularized,lei2015consistency}, spherical coordinates with a
Gaussian mixture \citep{sannapassino2022spectral}, and SCORE
\citep{jin2015fast}. \Cref{tab:pipelines} reports total error. The first two
slide along the fiber where the minority sits on the steep ray, by up to
$58\%$ of the arc, with the minority weight inflated $4$--$13\times$. SCORE
avoids the slide there but fails bimodally at LO, near the fiber's endpoint:
at $C=0.5$, $35\%$ of its replicates land about $105\%$ of the arc away.
Adding P1's small-radius filter to SCORE restores the slide. In a factorial of
features (ratio, angle, unit vector), clusterers ($k$-means, GMM) and the
filter, no variant keeps total error below $5\%$ of the arc at every cell
(\cref{app:pipelines}). Three predictions we registered about which component
matters failed (\cref{app:negative}), so we claim the pattern, not a
mechanism.

\begin{table}[t]
\caption{Median total error $|\hat\theta-\theta_0|$, \% of arc, $R=200$
(mean degree in parentheses), and fitted-to-true minority weight at HI. SCORE
at LO is bimodal. $^\dagger$Along-fiber RMSE over all $R=20$ chains.}
\label{tab:pipelines}\centering\small\setlength{\tabcolsep}{3.5pt}
\begin{tabular}{lrrrr}
\toprule
& HI & $w_{\min}{=}0.03$ & LO & $\hat w/w$\\
method & (720) & (500) & (500) & HI\\
\midrule
P1-cluster & $31.0$ & $49.9$ & $4.5$ & $4.4$\\
row-normalized SC & $43.5$ & $58.0$ & $3.3$ & $7.3$\\
spherical + GMM & $26.7$ & $37.4$ & $5.5$ & $4.0$\\
SCORE & $4.3$ & $7.4$ & $46.4$ & $2.0$\\
SCORE + filter & $26.7$ & $52.3$ & $3.7$ & $3.9$\\
P1-Prony & $10.2$ & $12.1$ & $48.7$ & $2.5$\\
\midrule
RayEM & $0.26$ & $0.74$ & $0.32$ & $0.99$\\
Bayesian LSBM & $0.81^\dagger$ & $2.3^\dagger$ & -- & $0.82$\\
\bottomrule
\end{tabular}
\end{table}

\paragraph{What avoids it.} Model-based fits, which are given the ray family and
keep the embedding's noise model, are accurate at every cell. \emph{RayEM} is a ray-constrained instance of Gaussian
mixture modeling of ASE rows under their central-limit covariance
\citep{athreya2016limit,rubindelanchy2022statistical}, fitted by EM
(\cref{app:rayem}). Its total error is
below $1\%$ of the arc everywhere tested, and the fitted minority weight is
unbiased. The LSBM authors' Bayesian sampler \citep{sannapassino2022latent},
with curves restricted to lines through the origin, lands within $0.3\%$ and
$1.8\%$ of the arc at the two cells we ran. RayEM's objective is not flat along
the fiber: its fit beats the truth by a median of $0.9$--$1.6$ units, the size
ordinary noise produces ($\tfrac12\chi^2_3\approx1.2$). The same embedding,
read through two kinds of statistic, is displaced along the fiber or not; the
displacement is a property of the summary, not of the embedding. We have not
separated the contribution of the noise model from that of the ray family.

\section{Regime Map and Practitioner Guidance}\label{sec:regime}\label{sec:practitioner}
\begin{figure*}[t]
\centering
\includegraphics[width=0.86\textwidth]{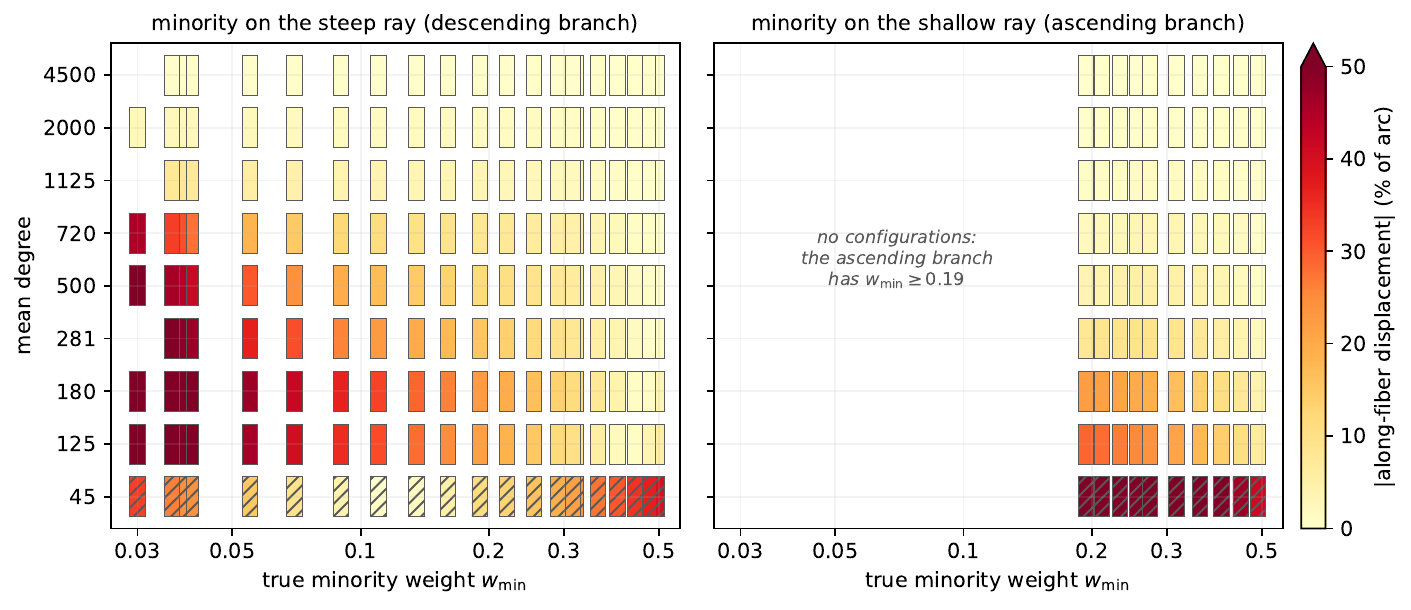}
\caption{Intrinsic along-fiber displacement of P1-cluster over $276$ cells
($9$ density levels; up to $21$ steep-ray and $10$ shallow-ray
configurations). Left: minority on the steep ray; right: on the shallow ray,
where $w_{\min}\ge0.19$. The safe region is top right. The hatched row (mean
degree $45$) is estimator breakdown (off-curve error).}
\label{fig:phase}
\end{figure*}

Displacement rises monotonically as the minority weight falls and fades with
density (\cref{fig:phase}): at $C=0.3$, $n=32{,}000$ it grows from $0.19\%$
at $w_{\min}=0.495$ to $45.5\%$ at $w_{\min}=0.03$. Three regimes appear. For
$n$ too small the error is large, off the curve, and roughly isotropic; the
estimator is visibly bad. In an intermediate regime the error collapses onto
the fiber, with at least $80\%$ of replicates within $0.02$ of it, and is
displaced by up to $54\%$ of the arc at $5$--$14$ times its spread. For $n$
large enough the displacement clears. At HI it runs
$48.8\%\to7.5\%\to0.24\%$ as $n$ grows from $2000$ to $32{,}000$ at
$C=0.75$, and $55.4\%\to44.6\%\to2.6\%$ at $C=0.5$; at $C=0.3$ it has not
cleared by $n=32{,}000$ ($31.4\%$). The clearing threshold rises as density
falls.

\paragraph{Check 1: am I in the regime?}
Look up displacement against mean degree, fitted minority weight, and which
ray carries the minority. Because the fitted minority weight is itself
inflated ($1.11\times$ at balanced weights, $4.67\times$ at pair B),
thresholds read off the true weight under-warn, and we state them on the
fitted scale (\cref{tab:safe} in \cref{app:tables}; $K=4$, uniform radial law). Displacement stays under $5\%$ at
every weight at mean degree $2000$ and above; with the minority on the steep
ray, a fitted weight of at least $0.095$ is needed at mean degree $1125$,
$0.37$ at $720$, and $0.42$ at $500$.

\paragraph{Check 2: what to run instead.}
Fit the embedding with a model that keeps its noise geometry rather than
summarizing each node by a direction. RayEM and the Bayesian LSBM are accurate
to within $1\%$ and $2.3\%$ of the arc at every cell tested, including the
cells where every summary pipeline fails (\cref{tab:pipelines}). They need $K$
and the ray family; their behavior when these are wrong is untested
(\cref{sec:scope}).

\paragraph{Check 3: why my confidence region is silent.}
Nominal $95\%$ regions built from sampling variability miss the truth at
every configuration in $17$ of $18$ (estimator, $C$, $n$) groups, and contain it
at $5.1\%$ in the best, even using the true replicate covariance. The error is bias in
every direction; the across-fiber offset alone is $1.5$--$33$ times its own
spread. Bootstrap, jackknife, sandwich and profile methods all estimate
variance and assume asymptotic unbiasedness, so none can bridge it. Nor does
the natural single-graph remedy: subsampling one graph and extrapolating in
$n$ flags low density rather than displacement, and Richardson-correcting the
estimates overshoots where the bias is small (\cref{app:negative}).

\paragraph{Check 4: how wide is the ambiguity?}
The fiber of \cref{thm:main} through the fit's own $(\hat m_1,\hat m_2)$
needs no solver at $K=4$ and shows how much the low-order summary leaves open: a fit giving
minority weight $0.17$ may sit on a fiber spanning $0.03$ to $0.50$. Read it
as a width, not a confidence set; it contains the truth reliably only in the
densest cells (\cref{tab:ambiguity} in \cref{app:tables}).

\section{Scope and Discussion}\label{sec:scope}
The shared radial law is load-bearing: it factors out of every density
(\cref{lem:reduction}) and reduces the census to a truncated moment problem.
The fiber nonetheless transfers upward: configurations agreeing below order
$2K-1$ still agree in the larger family where each ray has its own radial
scale, since enlarging a model class can only add non-identifiability, and a
practitioner cannot impose the restriction on data. What does not transfer is
the closed form; relaxing the symmetry adds parameters and census constraints,
and which grows faster is unproved.

\paragraph{Limitations.} The theory covers the symmetric family with a
shared radial law; the spectral experiments are at $K=4$, $n\le32{,}000$, on
one fiber, on simulated graphs. The census cost $n^*$ is extrapolated
(\cref{tab:census}). The message-passing result is a graphon-limit statement
for networks without node features; spectral positional encodings do carry the
slopes. Every estimator is given $K$, the model-based fits also the ray
family; we do not study selecting $K$, noting only that the census cannot
separate $K$ from $K-1$ at the fiber's open end, where two rays merge or the
middle ray empties (\cref{thm:xi_admissibility,thm:xi_admissibility_odd}).
Higher-order entrywise bias correction \citep{xie2024higher} might reduce the
tilt of summary pipelines; we have not tested it.

\section*{Use of Generative AI}

In this work, we used generative AI tools for formulating precise mathematical claims consistent with the standard terminology used in the literature, writing the proof and double-checking notations and consistencies in the proof, identifying gaps in the proof, designing and providing feedback for the suite of experiments as well as suggesting additional ablations, helping analyze the results and identifying bugs in the code that produced such results, assisting with the interpretation of results (final interpretations are the authors'), writing the test drivers and plotting routines, implementing the comparison estimators (RayEM and the published spectral pipelines) from author-approved designs, and identifying typos and non-material algebraic errors in the proof.

We have not used generative AI tools for supplying the main mathematical ideas of the proof (the main theorem and its key mathematical ideas belong to the authors; some secondary results, Corollaries~\ref{cor:mpnn} and~\ref{cor:census_char}, Proposition~\ref{prop:bound} and Corollary~\ref{cor:uniqueness_odd}, were proposed or proved with AI assistance and verified by the authors), or for main coding tasks such as the inference engine. Tasks such as cleaning and reformatting datasets and assisting with translation are not applicable to this work.

We also used generative AI tools to create plotting and graphing code from experimental results, suggest the suites of experiments and the range of parameters to run them over, draft major parts of the paper, correct grammar and style, search the literature for related work, summarize related papers, compare our paper with them so as not to overclaim novelty, and brainstorm further experiments in light of that work.

We have reviewed all AI-assisted work and take responsibility for the final content of this work, including text, claims, or artifacts produced with the aid of generative AI.

\bibliographystyle{plainnat}
\bibliography{refs}

\clearpage
\appendix

\section{Proofs}\label{app:proofs}
\subsection{Moments and motif density}
\label{sec:moments_motifs}

We call two graphs that are generated from two different configurations ``similar'' if the expected density of many subgraphs are the same. This is a loose term, but in the next few sections we will revisit more precisely which subgraph densities are matched and the smallest subgraph whose density can tell them apart, along with the characteristics of these configurations. For now, we would like to establish the link between subgraph densities and slope moments.

The link is developed in Sections~\ref{sec:hom_density}--\ref{sec:slope_index}. The density of a motif $F$ factors into a radial part, which depends only on $R$, and a slope part (Lemma~\ref{lem:factorization}). The slope part is a polynomial with nonnegative coefficients in the even slope moments $m_1, m_2, \dots$, indexed by the even subgraphs of $F$ (Theorem~\ref{thm:even_expansion}). The highest moment index that appears is a purely combinatorial invariant of $F$, its \emph{slope index} $h(F)$ (Definition~\ref{def:slope_index}). Consequently, configurations that share $R$ and match slope moments up to $m_j$ have identical densities for every motif on at most $2j+2$ vertices, and if they differ first at $m_{j+1}$ they are separated on $2j+3$ vertices, first by the friendship graph $F_{j+1}$ (Corollary~\ref{cor:census_transfer}).

\subsubsection{Slope Moments}
So now the immediate problem is, given a configuration, what other configurations can match its lower moments, and at which higher moments are they broken?

A few consequences of the generative model are immediate. 
\begin{proposition}
\label{prop:odd_vanish}
    Let $\mu_j$ be the $j$-th moment of the slope: $\mu_j \coloneq \mathbb{E}[s^j]$. Then $\mu_j = 0$ if $j$ is odd and positive if $j$ is even. We write $m_j \coloneq \mu_{2j}$.
\end{proposition}
\begin{proof}
    The slope law is symmetric, since $s_i = -s_{K+1-i}$ and $w_i = w_{K+1-i}$, so the odd moments vanish. For even $j$, $\mu_j = \sum_i w_i s_i^j > 0$ because the weights are positive and some slope is nonzero.
\end{proof}

\begin{definition}
\label{def:moment_matrix}
Let $\mathbb{M}_j$ be a $j \times j$ matrix that contains the even moments of the slope (or, moments of the squared slope). That is, $\mathbb{M}_{a,b} = m_{(a-1) + (b-1)}, 1 \leq a,b \leq j$. So the top-left corner of $\mathbb{M}_j$ is always 1. Let $D_j \coloneq \mathrm{det}(\mathbb{M}_j)$.
\end{definition}

We can see that $\mathbb{M}$ is constant along any anti-diagonal and contains the moments of $s$ from zero all the way up to $m_{2j-2}$ (the $4j-4$-th moment). Given two configuration $\Theta_1, \Theta_2$ we would like to characterize how much of the entries of $\mathbb{M}$ are matching.

\begin{definition}[Perfect matching of order $j$]
\label{def:perfect_matching}
    Configurations $\Theta_1, \Theta_2$ are perfectly matching of order $j$, written $\Theta_1 \sim_j \Theta_2$, if
    $$m_d(\Theta_1) = m_d(\Theta_2) \quad \forall 0 \leq d \leq j$$
    Equivalently $\mu_e(\Theta_1) = \mu_e(\Theta_2)$ for all $e \leq 2j$
\end{definition}

\begin{definition}[Near perfect matching of order $j$]
\label{def:near_perfect_matching}
    Configurations $\Theta_1, \Theta_2$ are nearly perfectly matching of order $j$, written $\Theta_1 \sim^o_j \Theta_2$, if all the equalities of Definition \ref{def:perfect_matching} hold, except possibly the one for the top monomial $m_j = \mu_{2j}$.
\end{definition}

Note that near perfect matching does not require the $m_j$ to be different, it just does not require them to be the same. Therefore, both $\sim_j$ and $\sim^o_j$ are equivalence relations (reflexive, symmetric, and transitive), and $\sim_j$ implies $\sim^o_j$.

\begin{definition}[Matching Order]
    Let $\mathrm{ord}(\Theta_1, \Theta_2) \coloneq \max\{ j : \Theta_1 \sim_j \Theta_2 \}$. That is, the maximum order such that both configurations match perfectly.
\end{definition}

\subsubsection{Homomorphism densities of the model}
\label{sec:hom_density}

Throughout, $F = (V(F), E(F))$ is a finite simple graph and $\deg_F(v)$ is the degree of $v$ in $F$. Write $\pi_\Theta \coloneq \sum_{i} w_i \delta_{s_i}$ for the \emph{slope law} of $\Theta$, so that $\mu_j = \mathbb{E}_{s \sim \pi_\Theta}[s^j]$. Write
$$\rho_d \coloneq \mathbb{E}_{x \sim R}[x^d], \quad d \geq 0$$
for the \emph{radial moments}. Since $R$ is supported on $(0,C)$, every $\rho_d$ is finite and positive.

The model of Section~\ref{sec:generative} is a graphon in the sense of Lov\'asz \citep{lovasz2012large}%
. Its latent space is $\Omega = [K] \times (0,C)$ with probability measure $P = \pi_\Theta \otimes R$, where we identify community $c$ with slope $s_c$, and its kernel is
$$W\big((c,x),(c',x')\big) = x x' (1 + s_c s_{c'}).$$

\begin{definition}[Homomorphism density]
    The homomorphism density of $F$ under $\Theta$ is
    $$t(F;\Theta) \coloneq \int_{\Omega^{V(F)}} \prod_{uv \in E(F)} W(z_u, z_v) \prod_{v \in V(F)} dP(z_v).$$
\end{definition}

This is the population quantity that motif counts estimate. Let $G_n$ be a graph on $n \geq |V(F)|$ nodes generated from $\Theta$, and let $t_{\mathrm{inj}}(F, G_n)$ be the number of injective homomorphisms $F \to G_n$ divided by $n(n-1)\cdots(n-|V(F)|+1)$.

\begin{proposition}[Motif counts are unbiased for $t$] \status{proved}
\label{prop:unbiased}
    $\mathbb{E}\left[ t_{\mathrm{inj}}(F, G_n) \right] = t(F;\Theta)$.
\end{proposition}
\begin{proof}
    Fix an injective map $\phi : V(F) \to [n]$. The latent vectors $(c_{\phi(v)}, x_{\phi(v)})$, $v \in V(F)$, are i.i.d.\ with law $P$, because $\phi$ is injective. Conditional on the latents, the edges $\phi(u)\phi(v)$, $uv \in E(F)$, are distinct node pairs and therefore independent, with probabilities $W$. So the probability that $\phi$ is a homomorphism is $t(F;\Theta)$. Average over $\phi$.
\end{proof}

Concentration of $t_{\mathrm{inj}}(F, G_n)$ around its mean as $n \to \infty$ is standard \citep{lovasz2012large}%
. We therefore identify the ``census'' with the population densities.

\begin{definition}[Census of order $m$]
    $\mathrm{Cen}_m(\Theta) \coloneq \big( t(F;\Theta) \big)_{F :\, |V(F)| \leq m}$.
\end{definition}

\begin{remark}
    Induced (graphlet) densities carry the same information as homomorphism densities at each order. For $F$ on vertex set $V$ with $|V| = m$, the injective density is a sum of induced densities, $t_{\mathrm{inj}}(F) = \sum_{F' \supseteq F} t_{\mathrm{ind}}(F')$, over graphs $F'$ on the same vertex set $V$ containing $F$. M\"obius inversion gives $t_{\mathrm{ind}}(F) = \sum_{F' \supseteq F} (-1)^{|E(F')| - |E(F)|} t_{\mathrm{inj}}(F')$. Hence two configurations agree on all induced densities of order $m$ if and only if they agree on all homomorphism densities of graphs on exactly $m$ vertices. Graphlet counts (e.g.\ ORCA) are thus covered by $\mathrm{Cen}_m$.
\end{remark}

\begin{lemma}[Radial--slope factorization] \status{proved}
\label{lem:factorization}
    $$t(F;\Theta) = \Bigg( \prod_{v \in V(F)} \rho_{\deg_F(v)} \Bigg) \cdot \tau(F; \pi_\Theta), \qquad \tau(F;\pi) \coloneq \mathbb{E}\Bigg[ \prod_{uv \in E(F)} (1 + s_u s_v) \Bigg],$$
    where $(s_v)_{v \in V(F)}$ are i.i.d.\ with law $\pi$.
\end{lemma}
\begin{proof}
    $\prod_{uv \in E(F)} x_u x_v = \prod_{v} x_v^{\deg_F(v)}$. The radii are independent of the slopes and of each other, so the expectation splits.
\end{proof}

\begin{remark}
    The fiber constructions of Sections~\ref{sec:construction_even} and~\ref{sec:construction_odd} vary only $(w, s)$ and keep $R$ fixed. Since every $\rho_d > 0$, two configurations sharing $R$ have $t(F;\Theta_1) = t(F;\Theta_2)$ if and only if $\tau(F;\pi_{\Theta_1}) = \tau(F;\pi_{\Theta_2})$, and the signs of the differences agree. $R$ enters the slope problem only through the admissibility threshold $T$ (Section~\ref{sec:range_xi}).
\end{remark}

\subsubsection{Even-subgraph expansion}
\label{sec:even_expansion}

\begin{definition}[Even subgraphs]
    A set $S \subseteq E(F)$ is \emph{even} if every vertex of $F$ has even degree in the spanning subgraph $(V(F), S)$. Write $\deg_S(v)$ for that degree and $\mathcal{E}(F)$ for the set of even subsets. $\mathcal{E}(F)$ contains $\emptyset$ and is the cycle space of $F$ over $\mathrm{GF}(2)$.
\end{definition}

\begin{theorem}[Even-subgraph expansion] \status{proved}
\label{thm:even_expansion}
    $$\tau(F;\pi_\Theta) = \sum_{S \in \mathcal{E}(F)} \ \prod_{v \in V(F)} m_{\deg_S(v)/2}.$$
\end{theorem}
\begin{proof}
    Expand the product: $\prod_{uv \in E(F)} (1 + s_u s_v) = \sum_{S \subseteq E(F)} \prod_{uv \in S} s_u s_v = \sum_{S \subseteq E(F)} \prod_{v} s_v^{\deg_S(v)}$. Take expectations using independence: the term for $S$ equals $\prod_v \mu_{\deg_S(v)}$. By Proposition \ref{prop:odd_vanish} (odd slope moments vanish), the term is zero unless every $\deg_S(v)$ is even, i.e.\ $S \in \mathcal{E}(F)$. For even $S$, $\mu_{\deg_S(v)} = m_{\deg_S(v)/2}$, and $m_0 = 1$ for vertices not touched by $S$.
\end{proof}

\begin{example}
    Forests have $\mathcal{E}(F) = \{\emptyset\}$, so $\tau = 1$ and $t(F;\Theta) = \prod_v \rho_{\deg_F(v)}$: tree densities (edge density, degree distribution, stars, paths) depend on $R$ only and are blind to the slopes. The cycle $C_\ell$ has $\mathcal{E} = \{\emptyset, E(C_\ell)\}$, so $\tau(C_\ell) = 1 + m_1^\ell$; in particular $t(\text{triangle}) = \rho_2^3 (1 + m_1^3)$.
\end{example}

\begin{definition}[Slope index]
\label{def:slope_index}
    $$h(F) \coloneq \max_{S \in \mathcal{E}(F)} \ \max_{v \in V(F)} \ \frac{\deg_S(v)}{2}.$$
    So $h(F) = 0$ if and only if $F$ is a forest.
\end{definition}

By Theorem~\ref{thm:even_expansion}, $\tau(F;\cdot)$ is a polynomial with nonnegative integer coefficients in $m_1, \dots, m_{h(F)}$, and it genuinely involves $m_{h(F)}$ (the maximizing $S$ contributes a monomial containing it).

\subsubsection{Moment matching and motif matching}
\label{sec:motif_transfer}

\begin{theorem}[Motif transfer] \status{proved}
\label{thm:motif_transfer}
    Let $\Theta_1, \Theta_2$ share the radial law $R$, and let $j \geq 0$.
    \begin{enumerate}
        \item If $\Theta_1 \sim_j \Theta_2$, then $t(F;\Theta_1) = t(F;\Theta_2)$ for every $F$ with $h(F) \leq j$.
        \item If $\Theta_1 \sim^o_j \Theta_2$ and $m_j(\Theta_1) > m_j(\Theta_2)$, then $t(F;\Theta_1) > t(F;\Theta_2)$ for every $F$ with $h(F) = j$.
    \end{enumerate}
\end{theorem}
\begin{proof}
    (1) By Theorem~\ref{thm:even_expansion}, $\tau(F;\cdot)$ depends only on $m_0, \dots, m_{h(F)}$, all of which agree. Apply Lemma~\ref{lem:factorization}.

    (2) Each term of Theorem~\ref{thm:even_expansion} has the form $m_j^{p_S} \cdot c_S$, where $p_S \geq 0$ counts vertices with $\deg_S(v) = 2j$ and $c_S > 0$ is a product of the matched moments $m_0, \dots, m_{j-1}$ (all positive by Proposition \ref{prop:odd_vanish}). Terms with $p_S = 0$ agree. Terms with $p_S \geq 1$ satisfy $m_j(\Theta_1)^{p_S} c_S > m_j(\Theta_2)^{p_S} c_S$ because $m_j > 0$ under both. At least one such term exists because $h(F) = j$. The radial prefactor is common and positive.
\end{proof}

\begin{remark}
    Part (2) is why near perfect matching $\sim^o_j$ is the natural relation: the separation needs only the \emph{order} of the first unmatched moment, not its value, and there is no cancellation. For $h(F) > j$, $t(F)$ also involves $m_{j+1}, m_{j+2}, \dots$, which are unconstrained by $\sim^o_j$; whether such $F$ separate every pair is not implied by this theorem. \status{open}
\end{remark}

\subsubsection{Combinatorics of the slope index}
\label{sec:slope_index}

\begin{proposition} \status{proved}
\label{prop:slope_index_bounds}
    Let $\Delta(F)$ be the maximum degree of $F$.
    \begin{enumerate}
        \item $h(F) \leq \lfloor \Delta(F)/2 \rfloor$.
        \item If $h(F) \geq j \geq 1$ then $|V(F)| \geq 2j+1$.
    \end{enumerate}
\end{proposition}
\begin{proof}
    (1) $\deg_S(v) \leq \deg_F(v)$ and $\deg_S(v)$ is even. (2) A vertex with $\deg_S(v) \geq 2j$ has at least $2j$ distinct neighbours.
\end{proof}

Part (1) is stronger than a vertex count: under $\sim_j$ every graph of maximum degree at most $2j+1$ is matched, however many vertices it has.

\begin{example}[Friendship graph]
\label{ex:friendship}
    Let $F_r$ ($r \geq 1$) be $r$ triangles sharing one common vertex (the \emph{center}), so $|V(F_r)| = 2r+1$ and $\Delta(F_r) = 2r$. The blocks of $F_r$ are its triangles, and the cycle space is the direct sum of the cycle spaces of the blocks, so the even subsets are exactly the unions of $i$ of the $r$ triangles, $0 \leq i \leq r$. Such a union gives the center degree $2i$ and gives $2i$ outer vertices degree $2$. Hence
    $$t(F_r;\Theta) = \rho_{2r}\, \rho_2^{2r} \sum_{i=0}^{r} \binom{r}{i} m_1^{2i}\, m_i, \qquad h(F_r) = r.$$
    The top moment $m_r$ enters with coefficient $\rho_{2r} \rho_2^{2r} m_1^{2r}$.
\end{example}

\begin{proposition}[Graphs of minimal order with $h(F) = j$] \status{proved}
\label{prop:minimal_order}
    Let $j \geq 1$ and $|V(F)| = 2j+1$. Then $h(F) = j$ if and only if $F$ has a vertex $v$ adjacent to all other vertices such that every connected component of $F[N(v)]$ (the subgraph induced on the neighbours of $v$) has an even number of vertices. $F_j$ is the edge-minimal such graph.
\end{proposition}
\begin{proof}
    ($\Rightarrow$) Take $S \in \mathcal{E}(F)$ and $v$ with $\deg_S(v) = 2j = |V(F)| - 1$. Then $v$ is adjacent (through $S$) to every other vertex. Let $S' = S \cap E(F[N(v)])$. For each $u \neq v$, $\deg_S(u) = 1 + \deg_{S'}(u)$ is even, so every vertex of $F[N(v)]$ has odd degree in $S'$. By the handshake lemma applied to each component, every component of $F[N(v)]$ has even order.

    ($\Leftarrow$) In each component of $F[N(v)]$, pair up its vertices arbitrarily and choose a path inside the component between each pair. Let $S'$ be the symmetric difference of the edge sets of all these paths. The parity of $\deg_{S'}(u)$ equals the number of paths having $u$ as an endpoint, which is one, so every vertex of $N(v)$ has odd degree in $S'$. Then $S = S' \cup \{vu : u \in N(v)\}$ is even with $\deg_S(v) = 2j$, so $h(F) \geq j$, and $h(F) \leq j$ by Proposition~\ref{prop:slope_index_bounds}.

    Edge-minimality: $v$ needs $2j$ edges, and $F[N(v)]$ needs at least $j$ edges (a graph on $2j$ vertices with no isolated vertices has at least $j$ edges). $F_j$ attains $3j$.
\end{proof}

\begin{corollary}[Census transfer] \status{proved}
\label{cor:census_transfer}
    Let $\Theta_1, \Theta_2$ share $R$, and let $j \geq 0$.
    \begin{enumerate}
        \item If $\Theta_1 \sim_j \Theta_2$ then $\mathrm{Cen}_{2j+2}(\Theta_1) = \mathrm{Cen}_{2j+2}(\Theta_2)$.
        \item If $\Theta_1 \sim^o_{j+1} \Theta_2$ and $m_{j+1}(\Theta_1) \neq m_{j+1}(\Theta_2)$ then $\mathrm{Cen}_{2j+3}(\Theta_1) \neq \mathrm{Cen}_{2j+3}(\Theta_2)$. Among graphs on exactly $2j+3$ vertices, the densities that differ are exactly those of the graphs characterized in Proposition~\ref{prop:minimal_order} (with $j+1$ in place of $j$), and each differs with the sign of $m_{j+1}(\Theta_1) - m_{j+1}(\Theta_2)$. For $j \geq 1$ the friendship graph gives
        $$t(F_{j+1};\Theta_1) - t(F_{j+1};\Theta_2) = \rho_{2j+2}\, \rho_2^{2j+2}\, m_1^{2j+2} \big( m_{j+1}(\Theta_1) - m_{j+1}(\Theta_2) \big).$$
    \end{enumerate}
\end{corollary}
\begin{proof}
    (1) If $|V(F)| \leq 2j+2$ then $h(F) \leq j$ by Proposition~\ref{prop:slope_index_bounds}(2); apply Theorem~\ref{thm:motif_transfer}(1).

    (2) A graph on $2j+3$ vertices has $h(F) \leq j+1$. If $h(F) \leq j$ its density agrees (Theorem~\ref{thm:motif_transfer}(1), since $\sim^o_{j+1}$ implies $\sim_j$). If $h(F) = j+1$ it differs, with the stated sign (Theorem~\ref{thm:motif_transfer}(2)). Proposition~\ref{prop:minimal_order} identifies the graphs with $h(F) = j+1$. For the displayed formula, use Example~\ref{ex:friendship}: every term with $i \leq j$ cancels, and $m_1$ is matched when $j \geq 1$.
\end{proof}

\begin{remark}
    Section~\ref{sec:fiber_motifs} applies Corollary~\ref{cor:census_transfer} with $j = K-2$ to the fibers constructed in Sections~\ref{sec:construction_even} and~\ref{sec:construction_odd}.
\end{remark}

\subsection{Configuration Construction for $K$ even}
\label{sec:construction_even}

Given a configuration $\Theta$, it is straightforward to compute the moments $m_j$ and the matrix $\mathbb{M}$. However, in this subsection we show the reverse. Given the moments $m_j$ up until a particular $j$, we show how to construct a configuration that realizes those moments (provided they are realizable). Furthermore, we show how to construct such a configuration from a ``seed'' real number $\xi$ which can vary. Therefore, we show how to construct a \emph{family} of configurations that all realize those moments. 

In outline: given a $\xi \in \mathbb{R}$, we would like to construct the matrix $J(\xi)$. The eigenvalues of $J(\xi)$ will show us how to construct the slopes $s_i$ and the eigenvectors of $J(\xi)$ will show us how to construct the weights $w_i$. Care must be taken so that the eigenvalues exist and are positive, that the community weights are positive and sum up to 1, and that the resulting dot products always land in $[0,1]$. These conditions will limit the range of $\xi$ we can choose, and will be addressed in the  admissibility analysis later on. For now, assume that we are given a range $[ \xi^-, \xi^+]$ such that the resulting matrix always has positive eigenvalues, and the resulting configuration will be admissible. 

We divide our analysis into even and odd $K$. In this section we first consider the case of $K$ even.

\begin{definition}[Construction of $\Theta$ for $K$ even]    
\label{def:construction_even}
Let $K = 2k$. Suppose $\xi \in [ \xi^-, \xi^+] \subset \mathbb{R}$, and let $m_0=1, m_1, \dots, m_{2k-2}$ be the desired moments. Call $\tilde{\nu}$ the probability measure that generated these moments (corresponding to the ``reference configuration'').

Let $\mathbb{M}_{k}$ be the moment matrix constructed as per Definition \ref{def:moment_matrix} (call this the Hankel matrix), and let $\mathbb{M}_k = U^T U$ be its Cholesky decomposition (so $U$ is $k \times k$). Note that $U$ is therefore upper triangular and positive diagonal. Set:
\begin{align*}
a_0 &= \frac{U_{1,2}}{U_{1,1}} \\
a_j &= \frac{U_{j+1,j+2}}{U_{j+1,j+1}} - \frac{U_{j,j+1}}{U_{j,j}} \quad 1 \leq j \leq k-2 \\
b_j &= \frac{U_{j+1,j+1}}{U_{j,j}}, \quad 1 \leq j \leq k-1 \\
J(\xi) &= \begin{bmatrix}
  a_0 & b_1 & & \\
  b_1 & a_1 & \ddots & \\ 
      & \ddots & \ddots & b_{k-1}\\ 
      &  & b_{k-1} & \xi \\ 
\end{bmatrix}
\end{align*}

Let $t_1 > \dots > t_k$ be the eigenvalues of $J(\xi)$. Assume $\xi^-, \xi^+$ are such that the resulting eigenvalues are all positive. Set $s_i = \sqrt{t_i}$ for $i \leq k$ and $s_i = -s_{K+1-i}$ for $i > k$ to be the slopes of the configuration.

Let $v_i$ be the corresponding orthonormal eigenvectors. Let $w_i = \frac{1}{2}(v_i)_1^2$ (half the squared first component of $v_i$) be the weights of the configuration for $i \leq k$, and $w_i = w_{K+1-i}$ for $i > k$.
When $k = 1$ there are no coefficients $a_j, b_j$ and $J(\xi) = [\xi]$.

Call $\nu_\xi$ the probability measure corresponding to this configuration. 
\end{definition}

The eigenvalues and eigenvectors of $J(\xi)$ can be found using numerical methods, but later in Section~\ref{sec:geometry}
 we will provide an explicit example for $K=4$ (or $k=2$). Now we shall prove that this construction achieves the desired moments. 

\subsubsection{Moment Matching Correctness}\label{sec:moment_matching_even}
First in the space of all polynomials with degree $\leq k-1$, let us define the inner product $\langle p, q \rangle = \mathbb{E}[ p(s^2) q(s^2) ]$. %
This inner product is well defined on that space, i.e.\ positive definite: the reference law of $s^2$ has exactly $k$ atoms, and a nonzero polynomial of degree at most $k-1$ cannot vanish at all of them. In particular $\mathbb{M}_k$ is positive definite, so the Cholesky decomposition used in Definition~\ref{def:construction_even} exists.

Note that because the space is limited to polynomials of degree $\leq k-1$, the expectations go up to the $m_{2k-2}$ which is given. In other words, these inner products exist, and can be assumed to be taken under the probability measure induced by the given moments. 

\begin{definition}[Polynomial Basis]
\label{def:polynomial_even}
Let $\langle p, q \rangle = \mathbb{E}[ p(s^2) q(s^2) ]$, and let $f_0, \dots, f_{k-1}$ be an orthogonal basis such that $f_j$ is monic and has degree exactly $j$. We can get such a basis by running Gram-Schmidt on $1,x,x^2,...,x^{k-1}$ for example. 
\end{definition}

This definition also means that $\langle x^i, x^j \rangle = m_{i+j}$. Also, orthogonality requires that $\langle f_i, f_j \rangle = 0$ for $i \neq j$. In other words, $f_j$ is orthogonal to any polynomial with degree less than $j$.

\begin{proposition}
\label{prop:f_even}
    The basis $f_j(x)$ as defined above must satisfy the following initial conditions and recurrence relation:
    $$f_0(x) = 1$$
    $$f_1(x) = x - m_1$$
    $$f_{j+1}(x) = (x-\tilde{a}_{j})f_j(x) - \tilde{b}_j^2f_{j-1}(x)$$
    for $j=1, \dots, k-2$ where 
    $$\tilde{a}_{j} = \frac{\langle xf_j, f_j \rangle}{|f_j|^2}$$
    $$\tilde{b}_j^2 = \frac{|f_j|^2}{|f_{j-1}|^2}$$
\end{proposition}

\begin{proof}
    $f_0(x) = 1$ is obvious. Since $f_1$ has to be a linear function, $f_1(x) = x - c$ for some $c$. But $\langle f_1, f_0 \rangle = m_1 - c = 0$ so $c = m_1$.

    Next, note that multiplication by $x$ is self-adjoint: $\langle xp, q \rangle = \mathbb{E}[s^2p(s^2)q(s^2)] = \langle p, xq \rangle$. 

    Now, $f_0, \dots, f_j$ are the orthogonal basis for polynomials of degree $\leq j$. Consider the polynomial $f_{j+1} - xf_j$. Both terms are monic polynomials of degree $j+1$, so their difference has degree at most $j$. Therefore there exist $c_0, \dots, c_j$ such that:
    $$f_{j+1} - xf_j = c_0 f_0 + \dots + c_j f_j$$

    When we apply inner product $\langle \cdot, f_i \rangle$ for $i = 0, \dots, j$. Note that $f_i$ is orthogonal to any polynomial with degree less than $i$, so there are very few survivors. In the LHS:
    $$\langle f_{j+1}-xf_j, f_i \rangle = - \langle xf_j, f_i \rangle = - \langle f_j, xf_i \rangle$$
    so the only survivors are $i = j-1$ (because $xf_i$ has degree $j$) and $i = j$ (because $xf_j$ has degree $j+1$). On the RHS, for a particular $i$, the only survivor is $c_i |f_i|^2$. Therefore:
    $$c_{j-1} = - \frac{\langle f_j, xf_{j-1}\rangle}{|f_{j-1}|^2}, c_j = - \frac{\langle xf_j, f_j\rangle}{|f_j|^2}$$
    We can see $c_j = -\tilde{a}_{j}$ readily. But for $c_{j-1}$, notice that $xf_{j-1}$ is again a monic polynomial of degree $j$, which can be expressed as $f_j$ plus a polynomial of degree less than $j$. Therefore $\langle f_j, xf_{j-1} \rangle = |f_j|^2$, and $c_{j-1} = -\tilde{b}_j^2$. The recurrence then holds.
\end{proof}

\begin{remark}
    Proposition \ref{prop:f_even} is a statement about uniqueness. We started with the assumption about inner product and orthogonality, and derived the recurrence. This is why the proposition asserts ``must satisfy''. Later we shall make use of this uniqueness result. 
\end{remark}

\begin{remark}
Note that $f_1$ also satisfies the recurrence formula if we let $\tilde{b}_0 f_{-1}$ terms as absent. There is nothing ``special'' about the case $j=0$.
\end{remark}

\begin{remark} 
The recurrence stops when $j=k-2$ at which point we get $f_{k-1}$ which is what we need as a basis for the space. That means $\tilde{a}_{k-2}$ is the highest order $\tilde{a}$ defined. But there is nothing stopping us from calculating $\tilde{a}_{k-1}$ since $f_{k-1}$ is well defined. In fact, all the previous $\tilde{a}$ requires computing $m_{2k-2}$, which are given (to be matched). So this is the first non-zero moment that lives outside the matched data, and this is the reason behind our main theorem as we shall later see. 
\end{remark}

\begin{remark} 
There is a nice interpretation for $\tilde{a}_{j} = \frac{\mathbb{E} \left[ xf_j(x)^2\right]}{\mathbb{E} \left[ f_j(x)^2\right]}$, the mean of $x$ under the reweighted measure of $f_j(x)^2$. The weights are non-negative, so $s_k^2 \leq \tilde{a}_{j} \leq s_1^2$ for all $j$. Later we shall see that this interpretation allows us to derive the range of $\xi$ that is admissible. 
\end{remark}

Now we shall prove that $\tilde{a}, \tilde{b}$ as defined in Proposition \ref{prop:f_even} is the same as the one in Definition \ref{def:construction_even}.
\begin{proposition}
\label{prop:a_b_equivalence}
Given the definitions of $\tilde{a}_j, a_j, \tilde{b}_j, b_j$ and the fact that $\mathbb{M}_{k} = U^T U $, we have:
    $$a_0 = m_1$$
    $$a_j = \tilde{a}_j  \quad 1 \leq j \leq k-2$$
    $$b_j = \tilde{b}_j  \quad 1 \leq j \leq k-1$$
\end{proposition}
\begin{proof}
    $\mathbb{M}_{a,b} = m_{a+b-2} = \mathbb{E}[s^{2(a+b-2)}] = \mathbb{E}[s^{2a-2} s^{2b-2}]= \langle x^{a-1}, x^{b-1} \rangle$. So $\mathbb{M}$ is the Gram matrix of the monomial basis $1, x, \dots, x^{k-1}$, which is not necessarily orthogonal. However, Cholesky decomposition is a unique decomposition that decomposes any basis into orthonormal basis components. 

    Define $p_i$ to be the scaled version of $f_i$ such that its magnitude is one.
    $$p_i \coloneq \frac{f_i}{|f_i|}$$

    Define $G$ to be a $k \times k$ matrix such that $G_{ij} \coloneq \langle x^{j-1}, p_{i-1} \rangle$. Note that because $f_i$ is orthogonal to any polynomial with degree less than $i$, then $G$ is upper triangular. The diagonal $G_{j,j} = \langle x^{j-1}, p_{j-1} \rangle = |f_{j-1}| > 0$. 

    Next, we note that $x^{a-1} = \sum_i G_{i,a}p_{i-1}$ by the definition of $G$. So $\mathbb{M}_{a,b} = \langle x^{a-1}, x^{b-1} \rangle = \sum_i G_{i,a} G_{i,b} = (G^TG)_{a,b}$, so then we must have $G = U$. Then:
    $$\tilde{b}_j = \frac{|f_j|}{|f_{j-1}|} = \frac{U_{j+1,j+1}}{U_{j, j}} = b_j$$

    Now, since $f_j$ is a monic polynomial of degree $j$, we can write:
    $$f_j = x^j + \gamma_{j} x^{j-1} + \dots$$
    (i.e. $\gamma_j$ is the subleading coefficient in $f_j$). Matching the coefficients in the recurrence in Proposition \ref{prop:f_even} versus the above, we have: $\tilde{a}_{j} = \gamma_j - \gamma_{j+1}$. 
    
    Now let us express $x^j$ in terms of the basis $p_0, \dots, p_j$:
    $$x^j = \sum_{i=0}^j \langle x^j, p_i\rangle p_i = \sum_{i=0}^j U_{i+1,j+1} \frac{f_i}{|f_i|} = \sum_{i=0}^j \frac{U_{i+1,j+1} f_i}{U_{i+1,i+1}}$$

    Now compare the coefficients of $x^{j-1}$. In the LHS it is zero. In the RHS, $f_j$ term contributes $ \frac{U_{j+1,j+1}}{U_{j+1,j+1}} \gamma_j = \gamma_j$. And $f_{j-1}$ term contributes $\frac{U_{j,j+1}}{U_{j,j}}$. So:

    $$\gamma_j = -\frac{U_{j,j+1}}{U_{j,j}} \quad 1 \leq j \leq k-1$$
    
    From this we get $a_j = \tilde{a}_j$ for $1 \leq j \leq k-2$. The only thing we need to prove now is the base case $a_0$. But that is evident because $f_1 = x-m_1$ so $\gamma_1 = -m_1$
\end{proof}

Now we have shown that most entries in the matrix $J(\xi)$ are indeed recurrence coefficients for an orthogonal monic basis for the moments of $s^2$, and gave an interpretation of those coefficients. Now we connect them to the parameters of the configuration $\Theta$.

\begin{proposition}[Structural Admissibility]
\label{prop:admissibility_even}
    The configuration $\Theta$ constructed in Definition \ref{def:construction_even} gives community weights that are positive and sum up to one, and that all eigenvalues (and consequently line slopes) are all distinct.
\end{proposition}

\begin{proof}
Let $V$ be the matrix of eigenvectors and $\Lambda$ be a diagonal matrix of eigenvalues, so that $J = V \Lambda V^{-1}$. Then let $J^d = V \Lambda^d V^{-1}$ for any $d$. Also recall that because $J$ is symmetric, then $V^T = V^{-1}$ so $J^d = V \Lambda^d V^{T}$. Then, under the probability measure $\nu_\xi$: 
$$\mathbb{E}_{\nu_\xi} \left[ s^{2d} \right] = \sum_{i=1}^{2k} w_i s_i^{2d} = 2 \sum_{i=1}^{k} w_i s_i^{2d} =  \sum_{i=1}^k (v_i)^2_1 t_i^d = (J^d)_{1,1}$$
For example when we set $d=0$ we get $m_0 = 1 = \sum w_i $ so we are guaranteed to get total community weights of 1.

Also because all $b_j > 0$ then $J$ is unreduced. No eigenvector has first entry zero. Indeed, write $(v)_i$ for the components of an eigenvector $v$. If $(v)_1 = 0$, then row 1 of $(J - \lambda I)v = 0$ gives $b_1 (v)_2 = 0$, so $(v)_2 = 0$. Then row 2 gives $b_2(v)_3 = 0$, and so on, so all components of $v$ are zero, i.e.\ $v = 0$, which is not an eigenvector. Also, the same process of reading the rows of $(J - \lambda I) v$ completely determines $v$, so the eigenvalues are all distinct and the eigenvectors all have nonzero first entry. Therefore this construction gives communities that are distinct, all have positive weights, and the weights sum up to 1. This is why we call this Structural Admissibility. Note that we have yet to prove that the eigenvalues are all positive.
\end{proof}

 Let $e_j$ be the coordinate vectors of the $j$-th basis (zero everywhere but 1 at $j$-th element). We have seen in the proof above the significance of $J^d$. Now we can apply $f$ to $J$.

 \begin{lemma}
 \label{lem:f_even}
 For all $0 \leq j \leq k-1$, if $f_j$ and $J$ are as defined in Definitions \ref{def:construction_even} and \ref{def:polynomial_even}, then:
 $$f_j(J) e_1 = |f_{j}| e_{j+1}$$
 \end{lemma}
 \begin{proof}
 
     For $j=0$, it is immediate.
     For $j=1$, $f_1(J)e_1 = (J - m_1 I) e_1$. The first element is $a_0 - m_1 = 0$, the second element is $b_1$, so this is $b_1e_2$. But $|f_1| = |f_0|b_1 = b_1$.

     Now suppose the claim holds for $j-1$ and $j$, where $1 \leq j \leq k-2$. We show it for $j+1$, using the recurrence in Proposition \ref{prop:f_even} (with $a_j = \tilde a_j$, $b_j = \tilde b_j$ by Proposition \ref{prop:a_b_equivalence}):
     $$f_{j+1}(J)e_1 = (J-a_jI)f_j(J)e_1 - b_j^2f_{j-1}(J)e_1$$
     $$=|f_j|\,J e_{j+1} - a_j|f_j|\,e_{j+1}- b_j^2 |f_{j-1}|\,e_j$$
     $$=|f_j|\,J e_{j+1} - a_j|f_j|\,e_{j+1}- b_j|f_j|\,e_j,$$
     since $b_j |f_{j-1}| = |f_j|$. We read off the $(j+1)$-th column from the definition of $J$:
     $$J e_{j+1} = b_j e_j + a_{j}e_{j+1} + b_{j+1}e_{j+2}$$
     So the $e_j$ and $e_{j+1}$ terms cancel out:
     $$f_{j+1}(J)e_1 = b_{j+1}|f_j|\,e_{j+2} = |f_{j+1}|\,e_{j+2}$$

     The range of indices is valid because $j+2 \leq k$, and at $j = k-2$, $b_{j+1} = b_{k-1}$ is defined.
 \end{proof}

 \begin{remark}
     Note that the proof above touches the columns $1, \dots, k-1$ of $J$, so only uses quantities that are known (given) from the desired moments, and did not involve $\xi$ at all. Furthermore, the proof of the lemma above is purely algebraic (cancellation of terms from the recurrence formula), so it holds regardless of $\xi$ and any admissibility.
 \end{remark}

The remark above is important because the definition of inner product (and therefore $|f_j|$) itself depends on the probability measure we are using. Up until now, all the inner products have been defined using the given moments $m_0, \dots, m_{2k-2}$, therefore using $\tilde{\nu}$. Now consider any $i,j$. If $i \neq j$ then $\langle f_i, f_j \rangle = 0$. But if $i = j$, the inner product is a scalar $|f_i|^2$. The proposition below shows that under $\nu_\xi$, the same result also holds.

\begin{proposition}
\label{prop:inner_matching}
    Under the probability measure $\nu_\xi$, we can define $\langle ., .\rangle_{\nu_\xi}$ the same way. Under this measure,
    $$\langle f_i, f_j \rangle_{\nu_\xi} = \delta_{ij} |f_i|^2$$
    where the norm on the RHS is taken under the original probability measure $\tilde{\nu}$
\end{proposition}
\begin{proof}

Take any polynomial $g(x) = \sum_d c_d x^d$. Now consider the moment $\mathbb{E}_{\nu_\xi} \left[ g(s^2) \right]$ under $\nu_\xi$. From the proof of Proposition \ref{prop:admissibility_even}, 
$$\mathbb{E}_{\nu_\xi} \left[ s^{2d} \right] = (J^d)_{1,1}$$
So,
$$\mathbb{E}_{\nu_\xi} \left[ g(s^2) \right] = \sum_d c_d (J^d)_{1,1} = g(J)_{1,1} = e_1^T g(J) e_1$$
Now apply it with $g(x) = f_i(x)f_j(x)$, and note that the inner product under $\nu_\xi$ is exactly the expected value of the product of these two polynomials.
$$\langle f_i, f_j \rangle_{\nu_\xi} = \mathbb{E}_{\nu_\xi} \left[ f_i(s^2) f_j(s^2) \right] = \mathbb{E}_{\nu_\xi} \left[ g(s^2) \right] = e_1^T g(J) e_1$$

But now because $J$ is symmetric, then powers of $J$ are symmetric too, so:
$$LHS = \left[ f_i(J) e_1\right]^T \left[ f_j(J) e_1\right]$$
And now apply Lemma \ref{lem:f_even} to both factors and multiply:
$$\mathrm{LHS} = |f_i|\,|f_j|\, e_{i+1}^\top e_{j+1} = \mathrm{RHS}$$
The lemma is a purely algebraic proof, no probability measure involved, as we have remarked. But all the norms are defined using the reference measure  $\tilde{\nu}$, so the RHS in this proposition is also defined using the reference measure. 
\end{proof}

Since $m_d = \langle x^a, x^b \rangle$ for any $a+b=d$, and we matched the inner product of the basis under both probability measures $\tilde{\nu}$ and $\nu_\xi$, then we also matched $m_d$ under both measures. Note that this applies to $a,b \leq k-1$ so for $d \leq 2k-2$, which are exactly the moments we are asked to match. In fact, $m_{2k-2}$ is the bottom-right corner of the matrix $\mathbb{M}_k$, the ``last'' element in the matrix that we needed to match.

However, we can push the computation one ``index'' higher and compute $m_{2k-1}$, or better yet, compute $\langle xf_{k-1}, f_{k-1} \rangle$ two ways, one of which will contain $\mathbb{E} \left[ x^{2k-1} \right]$ and the other way will draw from $J$

First, by definition of $a_{k-1}$:
$$\langle xf_{k-1}, f_{k-1} \rangle_{\nu_\xi} = a_{k-1}|f_{k-1}|^2$$
Second, applying Lemma \ref{lem:f_even}:
$$\langle xf_{k-1}, f_{k-1}  \rangle_{\nu_\xi} = (xf^2_{k-1})(J)_{1,1}= \left[f_{k-1}(J)e_1 \right]^T \cdot J \cdot \left[ f_{k-1}(J)e_1 \right]$$
$$= \left[|f_{k-1}|\,e_k \right]^T \cdot J \cdot \left[ |f_{k-1}|\,e_k \right] = |f_{k-1}|^2 J_{k,k} = \xi  |f_{k-1}|^2$$

So $a_{k-1} = \xi$. Note that this quantity contains $m_{2k-1}$ as its leading coefficient: expanding $\langle x f_{k-1}, f_{k-1} \rangle_{\nu_\xi} = m_{2k-1}(\nu_\xi) + (\text{terms in } m_0, \dots, m_{2k-2})$, and the lower terms are matched. So $m_{2k-1}(\nu_\xi) = |f_{k-1}|^2 \xi + c$ for a constant $c$ determined by the data, an affine function of $\xi$ with positive slope $|f_{k-1}|^2$. However, $\xi$ is precisely what generates the configuration $\Theta$. In other words, the configuration matches exactly the moments that were given to us, and not more. The first moment that we are not required to match is in fact free and varies by $\xi$.

\begin{corollary}[Moment matching and first unmatched moment] \status{proved}
\label{cor:first_unmatched_even}
    Suppose $\xi_1 \neq \xi_2$ generate $\Theta_1, \Theta_2$ with the construction defined in Definition \ref{def:construction_even}. Then the moments $m_0, \dots, m_{2k-2}$ of $\Theta_1, \Theta_2$ are equal, i.e.\ $\Theta_1 \sim^o_{2k-1} \Theta_2$. However, $m_{2k-1}$ are different. In fact, $m_{2k-1}(\nu_{\xi}) = |f_{k-1}|^2 \xi + c$ is strictly increasing and affine in $\xi$. In terms of $K = 2k$: the matched moments are $m_0, \dots, m_{K-2}$ and the first unmatched moment $m_{K-1}$ is strictly increasing in $\xi$.
\end{corollary}
\begin{proof} %
    The moments $m_0, \dots, m_{2k-2}$ match by Proposition \ref{prop:inner_matching} and the paragraph after it. For $m_{2k-1}$, the computation above gives $\langle xf_{k-1}, f_{k-1} \rangle_{\nu_\xi} = \xi |f_{k-1}|^2$. Expanding the left side, the coefficient of $m_{2k-1}(\nu_\xi)$ is one and every other term involves only $m_0, \dots, m_{2k-2}$, which are matched. So $m_{2k-1}(\nu_\xi) = |f_{k-1}|^2\xi + c$, and the slope $|f_{k-1}|^2$ is positive.
\end{proof}

\begin{corollary}[Uniqueness of construction]
\label{cor:uniqueness}
    Let $\Theta$ be any configuration with $K = 2k$ communities and probability measure $\nu_\Theta$, such that $m_d$ under $\nu_\Theta$ match for $0 \leq d \leq 2k-2$. Then $\nu_\Theta = \nu_\xi$ for some $\xi$. In other words, $\Theta$ must have been constructed using the procedure in Definition \ref{def:construction_even}
\end{corollary}
\begin{proof}
    Because the moments are the same up to $2k-2$, the $\mathbb{M}$ matrices are the same, and hence the factors $U$ are the same, and hence $f_j$ (due to uniqueness of monic orthogonal family). In fact, we simply apply the Gram--Schmidt process which is deterministic and well-defined. Then all the coefficients $a_j,b_j$ are the same too. The only free choice left is the corner $\xi = a_{k-1}$, and it is well defined because $\nu_\Theta$ is an actual measure so it has its own $m_{2k-1}$. %
    It remains to see that $\nu_\Theta$ is determined by this matrix. Since $\Theta$ has $K = 2k$ communities, $\nu_\Theta$ has exactly $k$ atoms, so it is the spectral measure at $e_1$ of its own $k \times k$ Jacobi matrix \citep[Section~1.3]{gautschi2004orthogonal}. That matrix is $J(\xi)$ with $\xi = \tilde a_{k-1}(\nu_\Theta)$, so $\nu_\Theta = \nu_\xi$ by the identity $\mathbb{E}_{\nu_\xi}[s^{2d}] = (J^d)_{1,1}$ in the proof of Proposition \ref{prop:admissibility_even}.
\end{proof}

The following proposition is not needed for the proof but it provides an insight into why the matrix $J$ is the way it is. The inspiration for this construction was taken from truncated Stieltjes moment problem. Note that by definition of $U$, it is a change of basis from $\mathbb{M}$'s native coordinate to its orthonormal coordinate, so in its orthonormal coordinate $\mathbb{M}$ is an identity matrix. Indeed $U^{-T} \mathbb{M} U^{-1} = I$. The following proposition shows that $J$ is the ``shifted-by-1'' moment matrix in the orthonormal coordinate. The shifting of moment matrix by 1 is the standard technique used in many moment problem solution. 

\begin{proposition}
\label{prop:insight}
    Let $\mathbb{M}^{(1)}$ be the shifted-by-1 moment matrix, that is $\mathbb{M}^{(1)}_{a,b} = m_{(a-1)+(b-1)+1} = m_{a+b-1}$ instead of $m_{a+b-2}$. Then
    $$U^{-T} \mathbb{M}^{(1)} U^{-1} = J(\xi_{\mathrm{ref}}), \qquad \xi_{\mathrm{ref}} = \tilde a_{k-1}(\tilde\nu)$$
\end{proposition}
\begin{proof}
    $\mathbb{M}^{(1)}_{a,b} = m_{a+b-1} = \langle x^{a-1}, x\cdot x^{b-1} \rangle $. Similar to the proof of Proposition \ref{prop:a_b_equivalence}, expand $x^{a-1} = \sum_i U_{i,a} p_{i-1}$ and $x\cdot x^{b-1} = \sum_j U_{j,b} x\,p_{j-1}$, substitute and use the linearity of the inner product, so
     $$\mathbb{M}^{(1)}_{a,b} = \sum_{i,j}U_{i,a}\langle p_{i-1}, x\,p_{j-1} \rangle  U_{j,b} = (U^T J U)_{a,b} $$
    The last equality can be gleaned from the definition of $J$. On the main diagonal, the entries are $a_{i-1} = \langle p_{i-1}, x\,p_{i-1} \rangle$. On the diagonal above the main diagonal, the entries are 
    $$\langle p_{i-1}, xp_i \rangle = \langle xp_{i-1}, p_i \rangle = \frac{\langle xf_{i-1}, f_i \rangle}{|f_{i-1}| |f_i|}$$ 
    But note that $xf_{i-1}$ is a monic polynomial with degree $i$, so it can be expressed as $f_i + ...$ where the rest of the terms are lower degree. So upon taking inner product with $f_i$, only the first term survives, so the entry becomes $\frac{\langle f_i, f_i \rangle}{|f_{i-1}| |f_i|} = b_i$. The entries below the main diagonal are proven the same way.
\end{proof}
\begin{remark}
    Two consequences from the proof of this proposition, both explain why $J$ has that particular structure. First, $J$ is tridiagonal because if $|i - j| > 1$ then nothing survives the inner product $\langle p_{i-1}, x\,p_{j-1} \rangle$.
    Second, all the entries of $\mathbb{M}$ have been determined by the data (or given to be matched), but when we shift by 1, the bottom-right corner of $\mathbb{M}^{(1)}$ is free. Because $U$ is upper triangular, transforming this matrix to and from $J$ means the bottom-right corner of $J$ remains free, and that is where we get to choose $\xi$.
    
    Imagine if an oracle gives us the next moment $m_{2k-1}$ instead, then one can apply this procedure to find a $\xi$ that would realize that moment. This $\xi$ always exists and the resulting configuration is structurally admissible. Whether or not it is admissible in the greater sense (dot products always in $[0,1]$ etc) is a different issue. The entire machinery in this subsection is algebraic and linear-algebraic. In a way, the fiber can be thought of as the list of configurations that realizes all possible $m_{2k-1}$ moments, truncated in order to guarantee admissibility. 
\end{remark}

\begin{remark}
    To summarize the construction, we are given Gram matrix $\mathbb{M}$, and $U$ is the basis change that makes it orthonormal in the space induced by the bilinear form. One can show that $J$ is the shifted Gram matrix in this coordinate system. The data can fill $\mathbb{M}$ completely, and it can fill the shifted $\mathbb{M}^{(1)}$ almost everywhere except its last corner. This freedom in the last corner is what generates the fiber. The matrix $J$ that we gave in closed form is merely this shifted matrix in the orthonormal coordinate. All the coefficients $a_j,b_j$ are merely entries of the basis change matrix $U$. These techniques are all known (Krein's canonical representations \citep{krein1977markov}; the Gauss--Radau rule as a modification of the last diagonal entry of the Jacobi matrix \citep{golub1973some}, with nodes and weights from its eigenpairs \citep{golub1969calculation}; and the one-parameter family of $k$-point rules of degree $2k-2$, whose node polynomials $f_k + c\,f_{k-1}$ are quasi-orthogonal \citep{shohat1937mechanical})%
. Our contribution is reducing the problem of matching submotif densities into truncated moment problem. 
\end{remark}

\subsubsection{Admissibility Analysis}
In Proposition \ref{prop:admissibility_even} we showed that for any $\xi$, the resulting lines are all distinct, the community weights are all positive and sum up to 1. However, we still need to prove 1) that different $\xi$ will give different configurations, 2) identify the range allowed so that the resulting dot product will be in the interval $[0,1]$, and 3) identify the range allowed so that the eigenvalues are positive.

\paragraph{Monotonicity}

\begin{proposition}[Monotone Transport] \status{proved}
\label{prop:monotone_transport}
    Each eigenvalue $t_i(\xi)$ is a continuously differentiable, strictly increasing function of $\xi$, with
    $$\frac{dt_i}{d\xi} = (v_i)_k^2 > 0, \qquad \sum_{i=1}^k \frac{dt_i}{d\xi} = 1.$$
    Consequently $\xi_1 - \xi_2 = \sum_{i=1}^k \left( t_i(\xi_1) - t_i(\xi_2) \right)$, and different $\xi$ give different configurations. Moreover $t_k(\xi) \leq \xi \leq t_1(\xi)$.
\end{proposition}
\begin{proof}
The eigenvalues are simple (Proposition \ref{prop:admissibility_even}), hence differentiable in $\xi$ with derivative given by the Hellmann--Feynman formula. Since $\frac{\partial}{\partial \xi}J = e_k e_k^T$,
$$\frac{dt_i}{d \xi} = v_i^T e_k e_k^T v_i = (v_i)_k^2.$$
This is nonzero: reading the rows of $(J - \lambda I)v = 0$ from the bottom, row $k$ gives $b_{k-1} v_{k-1} = (\lambda - \xi) v_k$, so $v_k = 0$ forces $v_{k-1} = 0$, then row $k-1$ forces $v_{k-2} = 0$, and so on, giving $v = 0$. The sum is $\sum_{i} (v_i)_k^2 = 1$, the squared norm of row $k$ of the orthogonal matrix $V$. (Equivalently, $\sum_i t_i = \mathrm{tr}\, J(\xi) = \sum_{j=0}^{k-2} a_j + \xi$.) Integrating gives the displayed identity. Finally, $t_k \leq e_k^T J e_k = \xi \leq t_1$ by the Rayleigh quotient.
\end{proof}

\paragraph{Range of $\xi$}
\label{sec:range_xi}

Recall that for each pair of nodes $u,v$ belonging to communities $i,j$, $W(u,v) = x_{1,u} x_{1,v} (1+s_i s_j)$ with $x_{1,u},x_{1,v} \in [0,C]$. We need that $ 0 \leq W \leq 1$. First, consider the condition that $W \geq 0$. The negative factor can come from $(1+s_i s_j)$ and the smallest value this can take is when $i=1, j = K$ (the outermost two lines). We have $1+s_is_j = 1 - s_1^2 = 1-t_1 \geq 0$ so we need $t_1 \leq 1$

Next consider the condition $W \leq 1$. The maximum value $W$ can achieve is when both points are sampled from the farthest possible distance radially, and that $i=j=1$, so $C^2(1+t_1) \leq 1$ so $t_1 \leq 1/C^2 - 1$. So we need $t_1 \leq T = \min(1, 1/C^2 - 1)$. Since $R$ is supported on the open interval $(0,C)$, $W < 1$ strictly even when $t_1 = T$, so the endpoint $t_1 = T$ is admissible. Now we will look for the range of $\xi$ such that all eigenvalues are positive and are no greater than $T$.

\begin{lemma}[Characteristic polynomials] \status{proved}
\label{lem:charpoly}
    Let $J_j$ be the leading $j \times j$ block of $J(\xi)$ (independent of $\xi$ for $j \leq k-1$). Then $f_j(x) = \det(xI - J_j)$ for $1 \leq j \leq k-1$, and
    $$P_\xi(x) \coloneq \det(xI - J(\xi)) = (x - \xi) f_{k-1}(x) - b_{k-1}^2 f_{k-2}(x),$$
    with the conventions $f_{-1} \coloneq 0$ and $b_0 \coloneq 0$ (so $P_\xi(x) = x - \xi$ when $k = 1$).
\end{lemma}
\begin{proof}
    Expanding $\det(xI - J_{j+1})$ along its last row gives $\det(xI - J_{j+1}) = (x - a_j)\det(xI - J_j) - b_j^2 \det(xI - J_{j-1})$, with $\det(xI - J_0) = 1$ and $\det(xI - J_1) = x - a_0 = x - m_1$. This is the recurrence and initial condition of Proposition \ref{prop:f_even}, with $a_j = \tilde{a}_j$, $b_j = \tilde{b}_j$ by Proposition \ref{prop:a_b_equivalence}. The same expansion for $J(\xi)$, whose last diagonal entry is $\xi$, gives $P_\xi$.
\end{proof}

\begin{theorem} \status{proved}
    \label{thm:xi_admissibility}
    Let the reference configuration be admissible, let $T = \min \left(1, \frac{1}{C^2} - 1 \right)$, and let
    $$\xi^- = - \frac{b_{k-1}^2 f_{k-2}(0)}{f_{k-1}(0)}$$
    $$\xi^+ = T - \frac{b_{k-1}^2 f_{k-2}(T)}{f_{k-1}(T)}$$
    Then all eigenvalues of $J(\xi)$ lie in $(0, T]$ (so the configuration is admissible) if and only if $\xi \in (\xi^-, \xi^+]$. This interval is nonempty and contains the reference value $\xi_{\mathrm{ref}} = \tilde{a}_{k-1}(\tilde{\nu})$. These are the Gauss--Radau rules of $\tilde{\nu}$ with a prescribed node at $0$ and at $T$ \citep{golub1973some,gautschi2004orthogonal,golub2010matrices}%
. For $k = 1$, $\xi^- = 0$ and $\xi^+ = T$.
\end{theorem}

\begin{proof}
    
The case $k=1$ is immediate: $J(\xi) = [\xi]$, so $t_1 = \xi$ and the condition is $0 < \xi \leq T$. Assume $k \geq 2$.

\emph{Step 1 (interlacing).} Let $J_{k-1}$ be the leading $(k-1) \times (k-1)$ block, which does not depend on $\xi$. By Lemma \ref{lem:charpoly}, its eigenvalues $\theta_1 > \dots > \theta_{k-1}$ are the zeros of $f_{k-1}$. Cauchy interlacing gives, for every $\xi$:
$$t_1 > \theta_1 > t_2 > \theta_2 > \dots > \theta_{k-1} > t_k.$$
(Strict, because $J$ is unreduced.) By Corollary \ref{cor:uniqueness}, the reference measure is $\tilde{\nu} = \nu_{\xi_{\mathrm{ref}}}$ for some $\xi_{\mathrm{ref}}$, so the interlacing also holds for the reference eigenvalues $\tilde{t}_1 > \dots > \tilde{t}_k$. The reference is admissible, so $0 < \tilde{t}_k$ and $\tilde{t}_1 \leq T$, hence every $\theta_j \in (\tilde{t}_k, \tilde{t}_1) \subset (0, T)$. Therefore $t_2, \dots, t_{k-1} \in (\theta_{k-1}, \theta_1) \subset (0,T)$ for every $\xi$: only $t_1$ can exceed $T$ and only $t_k$ can be nonpositive.

\emph{Step 2 (monotonicity).} By Proposition \ref{prop:monotone_transport}, $t_k$ and $t_1$ are continuous and strictly increasing, with $t_k(\xi) \leq \xi \to -\infty$ as $\xi \to -\infty$ and $t_1(\xi) \geq \xi \to \infty$ as $\xi \to \infty$. Hence $\{\xi : t_k(\xi) > 0\} = (\xi^-, \infty)$ where $\xi^-$ is the unique solution of $t_k(\xi) = 0$, and $\{\xi : t_1(\xi) \leq T\} = (-\infty, \xi^+]$ where $\xi^+$ is the unique solution of $t_1(\xi) = T$. The admissible set is their intersection $(\xi^-, \xi^+]$.

\emph{Step 3 (closed form).} By Lemma \ref{lem:charpoly}, a number $a$ is an eigenvalue of $J(\xi)$ if and only if $P_\xi(a) = 0$. When $f_{k-1}(a) \neq 0$ this is equivalent to
$$\xi = \xi(a) \coloneq a - \frac{b_{k-1}^2 f_{k-2}(a)}{f_{k-1}(a)}.$$
This is well defined at $a = 0$ and $a = T$ because the zeros $\theta_j$ of $f_{k-1}$ lie in $(0,T)$. At $\xi = \xi(0)$, the eigenvalue $0$ must be $t_k$, since $t_1, \dots, t_{k-1} > \theta_{k-1} > 0$; so $\xi^- = \xi(0)$. At $\xi = \xi(T)$, the eigenvalue $T$ must be $t_1$, since $t_2, \dots, t_k < \theta_1 < T$; so $\xi^+ = \xi(T)$. This gives the stated formulas.

\emph{Step 4 (nonempty).} The reference is admissible, so $\xi_{\mathrm{ref}} \in (\xi^-, \xi^+]$. By the computation preceding Corollary \ref{cor:first_unmatched_even}, $\xi_{\mathrm{ref}} = \tilde{a}_{k-1}(\tilde{\nu})$.
\end{proof}

\subsection{Construction for $K$ odd}
\label{sec:construction_odd}

For $K=2k+1$, the symmetry condition requires that $s_{k+1} = 0$. Just like in the even $K$ case, we still would like to match $m_0 = 1, m_1, \dots, m_{K-2} = m_{2k-1}$. We will attempt a very similar construction as in Definition \ref{def:construction_even} and prove correctness and admissibility. The main insight still mimics that of Proposition \ref{prop:insight}, but we scale the entire matrix $\mathbb{M}$ (and therefore $\mathbb{M}^{(1)}$) so that the mean $m_1$ matches. The structural admissibility (Proposition \ref{prop:admissibility_even}) follows by construction.

\begin{definition}
\label{def:construction_odd}
    Set $m'_d = \frac{m_{d+1}}{m_1}, 0 \leq d \leq 2k-2$ (this uses $m_1, \dots, m_{2k-1}$). Apply Definition \ref{def:construction_even} to $(m'_0, \dots, m'_{2k-2})$. Denote the resulting coefficients, polynomials and matrix by $a'_j$, $b'_j$, $f'_j$ and $J'(\xi)$, its eigenvalues by $t'_1 > \dots > t'_k$ and its orthonormal eigenvectors by $v'_1, \dots, v'_k$. Let $\nu'_\xi \coloneq \sum_{i=1}^k (v'_i)_1^2\, \delta_{t'_i}$, the law of $s^2$ under that even construction. Assume $J'(\xi) \succ 0$. The slopes of the configuration are: $s_i = \sqrt{t'_i}$ for $i \leq k$, $s_{k+1} = 0$, and $s_{i} = -s_{K+1-i}$ for $i>k+1$.
    The community weights are $$w_i = \frac{m_1}{2t'_i} (v'_i)_1^2$$ 
    for $i \leq k$, $w_i = w_{K+1-i}$ for $i > k+1$, and 
    $$w_{k+1} = 1 - \sum_{i \neq k+1} w_i$$
    Call $\mu_\xi$ the law of $s^2$ under this configuration: $\mu_\xi = w_{k+1}\delta_0 + \sum_{i=1}^k 2 w_i\, \delta_{t'_i}$.
\end{definition}

We will prove moment matching and structural admissibility by simple scaling argument. The idea is that we construct a $K-1$ configuration that matches the ``scaled'' version of the moments, while leaving some wiggle room for the middle community, whose weight must then be determined in order to make the weights sum up to 1.

\begin{theorem} \status{proved}
\label{theorem:correctness_odd}
For every $\xi$ with $J'(\xi) \succ 0$, the configuration described in Definition \ref{def:construction_odd} matches the required moments $m_0, \dots, m_{2k-1}$, its outer weights $w_i$ ($i \neq k+1$) are positive, and its nonzero slopes are distinct. The middle weight $w_{k+1}$ is not automatically positive; see Section~\ref{sec:admissibility_odd}.
\end{theorem}
\begin{proof}
    For $d = 0$, $m_0(\mu_\xi) = 1$ by the definition of $w_{k+1}$. For $1 \leq d \leq 2k-1$, the middle community contributes $0^d = 0$, so
    $$m_{d}(\mu_{\xi}) = \sum_{i=1}^{k} 2 w_i (t'_i)^{d} = m_1 \sum_{i=1}^k (v'_i)_1^2 (t'_i)^{d-1} = m_1\, m'_{d-1}(\nu'_\xi).$$
    Since $d - 1 \leq 2k-2$, the even construction matches $m'_{d-1}(\nu'_\xi) = m'_{d-1}$ (Section~\ref{sec:moment_matching_even})%
, so $m_d(\mu_\xi) = m_1 m'_{d-1} = m_d$.

    Positivity of $t'_i$ (assumed) and of $(v'_i)_1^2$ (Proposition \ref{prop:admissibility_even}) gives $w_i > 0$ for $i \neq k+1$, and distinctness of the $t'_i$ gives distinct nonzero slopes.
\end{proof}

\begin{corollary}[First unmatched moment, $K$ odd] \status{proved}
\label{cor:first_unmatched_odd}
    $m_{2k}(\mu_\xi) = m_1\, m'_{2k-1}(\nu'_\xi) = m_1 \big( |f'_{k-1}|^2 \xi + c' \big)$ for a constant $c'$ determined by the data. In terms of $K = 2k+1$: the matched moments are $m_0, \dots, m_{K-2}$ and the first unmatched moment $m_{K-1}$ is strictly increasing and affine in $\xi$.
\end{corollary}
\begin{proof}
    The first equality is the computation above with $d = 2k$. The second is Corollary \ref{cor:first_unmatched_even} applied to the primed construction.
\end{proof}

\subsubsection{Admissibility Analysis for $K$ odd}
\label{sec:admissibility_odd}

The middle ray imposes no new constraint on $W$: within the middle community $W = x_u x_v \leq C^2 \leq 1$ (as $T > 0$ requires $C^2 < 1$), and between the middle and community $i$, $1 + s_i \cdot 0 = 1 \geq 0$. So, as in Section \ref{sec:range_xi}, the conditions are $0 < t'_k$, $t'_1 \leq T$, and in addition $w_{k+1} > 0$. We show that the last condition, not $t'_k > 0$, determines the lower endpoint.

\begin{lemma}[Middle weight] \status{proved}
\label{lem:middle_weight}
    If $J'(\xi) \succ 0$ then
    $$w_{k+1} = 1 - m_1\, g(\xi), \qquad g(\xi) \coloneq e_1^T J'(\xi)^{-1} e_1 = \sum_{i=1}^k \frac{(v'_i)_1^2}{t'_i}.$$
    So $w_{k+1} > 0$ if and only if $g(\xi) < 1/m_1$.
\end{lemma}
\begin{proof}
    $\sum_{i \neq k+1} w_i = 2 \sum_{i=1}^k \frac{m_1}{2 t'_i} (v'_i)_1^2 = m_1 \sum_i \frac{(v'_i)_1^2}{t'_i}$, and $J'^{-1} = \sum_i (t'_i)^{-1} v'_i v'^T_i$.
\end{proof}

\begin{proposition} \status{proved}
\label{prop:g_monotone}
    Let $\xi'_0 \coloneq -b'^2_{k-1} f'_{k-2}(0) / f'_{k-1}(0)$, so that $J'(\xi) \succ 0$ if and only if $\xi > \xi'_0$ (Theorem \ref{thm:xi_admissibility}, Steps 1--3, applied to the primed construction; its interlacing hypothesis holds by Step 4 of Theorem \ref{thm:xi_admissibility_odd} below). On $(\xi'_0, \infty)$, $g$ is strictly decreasing, and $g(\xi) \to +\infty$ as $\xi \downarrow \xi'_0$.
\end{proposition}
\begin{proof}
    Since $\frac{d}{d\xi} J' = e_k e_k^T$, we have $\frac{d}{d\xi} J'^{-1} = -J'^{-1} e_k e_k^T J'^{-1}$, so
    $$g'(\xi) = -\big( (J'^{-1})_{1k} \big)^2.$$
    By the cofactor formula, $(J'^{-1})_{1k} = (-1)^{k+1} \det(\text{minor}_{k,1}) / \det J'$. The minor obtained by deleting row $k$ and column $1$ of the tridiagonal $J'$ is lower triangular with diagonal $b'_1, \dots, b'_{k-1}$, so $(J'^{-1})_{1k} = (-1)^{k+1} \prod_{j=1}^{k-1} b'_j / \det J' \neq 0$. (For $k=1$, $g = 1/\xi$.) Hence $g' < 0$.

    For the limit: $g(\xi) \geq (v'_k)_1^2 / t'_k$. As $\xi \downarrow \xi'_0$, $t'_k \downarrow 0$ while $(v'_k)_1^2$ tends to a positive limit (it is continuous and nonzero at $\xi'_0$ by Proposition \ref{prop:admissibility_even}), so $g \to +\infty$.
\end{proof}

So $w_{k+1} \to -\infty$ as $t'_k \to 0^+$: the middle weight vanishes strictly before any slope does, and positivity of $t'_k$ never binds. The next result identifies where it vanishes.

\begin{theorem}[Admissible range, $K$ odd] \status{proved}
\label{thm:xi_admissibility_odd}
    Let $K = 2k+1 \geq 3$ and let the reference configuration be admissible. Let $\alpha_0, \dots, \alpha_{k-1}$ be the recurrence coefficients of Proposition \ref{prop:f_even} for the reference law of $s^2$ (moments $m_0, \dots, m_{2k-1}$, all given), and let $F_k$ be its degree-$k$ monic orthogonal polynomial. Set
    $$\xi^- = \sum_{j=0}^{k-1} \alpha_j - \sum_{j=0}^{k-2} a'_j, \qquad \xi^+ = T - \frac{b'^2_{k-1} f'_{k-2}(T)}{f'_{k-1}(T)},$$
    with the conventions of Lemma \ref{lem:charpoly} (so $\xi^+ = T$ when $k=1$). Then the configuration of Definition \ref{def:construction_odd} is admissible (all $t'_i \in (0,T]$ and $w_{k+1} > 0$) if and only if $\xi \in (\xi^-, \xi^+]$. The interval is nonempty and contains $\xi_{\mathrm{ref}}$. At $\xi = \xi^-$, $w_{k+1} = 0$ and the configuration is the $k$-point Gauss quadrature rule of the reference: the $2k$ outer rays sit at $\pm\sqrt{x}$ for the zeros $x$ of $F_k$, with the Christoffel numbers as weights.
\end{theorem}
\begin{proof}
    \emph{Step 1 (the admissible set is an interval with lower end $\xi^* > \xi'_0$).} Let $\xi_{\mathrm{ref}}$ be such that $\nu'_{\xi_{\mathrm{ref}}}$ is the $x$-reweighting of the reference law of $s^2$ (it exists by Corollary \ref{cor:uniqueness}, since that reweighting is a $k$-atom measure with moments $m'_0, \dots, m'_{2k-2}$). The reference has $w_{k+1} > 0$, so $g(\xi_{\mathrm{ref}}) < 1/m_1$. By Proposition \ref{prop:g_monotone} and the intermediate value theorem there is a unique $\xi^* \in (\xi'_0, \xi_{\mathrm{ref}})$ with $g(\xi^*) = 1/m_1$, and $\{\xi > \xi'_0 : w_{k+1} > 0\} = (\xi^*, \infty)$. On this set all $t'_i > 0$.

    \emph{Step 2 ($\xi^*$ is the Gauss rule).} The reference law of $s^2$ is supported on $k+1$ points ($0$ and $\tilde{t}_1, \dots, \tilde{t}_k$), so its $k \times k$ Hankel matrix is positive definite and $F_k$ is well defined from $m_0, \dots, m_{2k-1}$. The $k$-point Gauss rule $\gamma = \sum_{i=1}^k \lambda_i \delta_{x_i}$ has nodes at the zeros of $F_k$, positive Christoffel numbers $\lambda_i$, and is exact for polynomials of degree $\leq 2k-1$ \citep{gautschi2004orthogonal,szego1939orthogonal}%
. Its nodes lie in the interior of the convex hull of the reference support, $(0, \tilde{t}_1) \subset (0,T)$. Exactness gives $\sum_i \lambda_i x_i = m_1$, so $\gamma' \coloneq \sum_i \frac{\lambda_i x_i}{m_1} \delta_{x_i}$ is a $k$-atom probability measure with moments $m'_0, \dots, m'_{2k-2}$. %
    By Corollary \ref{cor:uniqueness}, $\gamma' = \nu'_{\hat\xi}$ for some $\hat\xi$, with $t'_i(\hat\xi) = x_i > 0$ and $(v'_i(\hat\xi))_1^2 = \lambda_i x_i / m_1$. Then
    $$w_{k+1}(\hat\xi) = 1 - m_1 \sum_i \frac{\lambda_i x_i / m_1}{x_i} = 1 - \sum_i \lambda_i = 0.$$
    By uniqueness in Step 1, $\hat\xi = \xi^*$. At this point $\mu_{\xi^*} = \gamma$, which is the Gauss rule claimed.

    \emph{Step 3 (closed form).} By Proposition \ref{prop:monotone_transport} (the trace identity) applied to $J'$, $\sum_i t'_i(\xi) = \sum_{j=0}^{k-2} a'_j + \xi$. At $\xi^*$ the left side is $\sum_i x_i$, the sum of the zeros of $F_k$, which is minus its subleading coefficient. From the recurrence of Proposition \ref{prop:f_even} for the reference, the subleading coefficient $\gamma_k$ of $F_k$ satisfies $\gamma_{j+1} = \gamma_j - \alpha_j$ with $\gamma_0 = 0$ (as in the proof of Proposition \ref{prop:a_b_equivalence}), so $\sum_i x_i = -\gamma_k = \sum_{j=0}^{k-1} \alpha_j$. Hence $\xi^* = \sum_{j=0}^{k-1} \alpha_j - \sum_{j=0}^{k-2} a'_j = \xi^-$.

    \emph{Step 4 (upper endpoint).} The middle ray adds no constraint on $t'_1$, so the condition $t'_1 \leq T$ is $\xi \leq \xi^+$ exactly as in Theorem \ref{thm:xi_admissibility}, applied to the primed construction. The interlacing step there needs the zeros of $f'_{k-1}$ to lie in $(0,T)$; this holds because $\nu'_{\xi_{\mathrm{ref}}}$ is supported on $\{\tilde{t}_1, \dots, \tilde{t}_k\} \subset (0, T]$.

    \emph{Step 5 (nonempty).} The reference is admissible, so $\xi_{\mathrm{ref}} \in (\xi^-, \xi^+]$.
\end{proof}

\begin{corollary}[Uniqueness of construction, $K$ odd] \status{proved}
\label{cor:uniqueness_odd}
    Let $K = 2k+1$, and let $\Theta$ be an admissible configuration with $K$ communities and radial law $R$ whose moments $m_0, \dots, m_{2k-1}$ match those of the reference. Then $\Theta$ is the configuration of Definition \ref{def:construction_odd} for some $\xi \in (\xi^-, \xi^+]$.
\end{corollary}
\begin{proof} %
    Write $\mu_\Theta = w_{k+1}\delta_0 + \sum_{i=1}^k 2w_i\,\delta_{t_i}$ with $t_1 > \dots > t_k > 0$. The atom at $0$ does not contribute to $m_d$ for $d \geq 1$, so $m_1 = \sum_i 2w_it_i > 0$ and
    $$\nu'_\Theta \coloneq \sum_{i=1}^k \frac{2w_it_i}{m_1}\,\delta_{t_i}$$
    is a probability measure with exactly $k$ atoms and moments $m_{d+1}/m_1 = m'_d$ for $0 \leq d \leq 2k-2$. By Corollary \ref{cor:uniqueness} applied to the primed data, $\nu'_\Theta = \nu'_\xi$ for $\xi = \tilde a'_{k-1}(\nu'_\Theta)$. So $t_i = t'_i$ and $2w_it_i/m_1 = (v'_i)_1^2$, i.e.\ $w_i = \frac{m_1}{2t'_i}(v'_i)_1^2$, the outer weights of Definition \ref{def:construction_odd}. The middle weight is $1 - \sum_{i \neq k+1} w_i$ in both. Finally, $\Theta$ is admissible, so $\xi \in (\xi^-, \xi^+]$ by Theorem \ref{thm:xi_admissibility_odd}.
\end{proof}

\begin{example}[$K = 3$]
    Here $k = 1$, $J' = [\xi]$, $t'_1 = \xi$, $w_1 = w_3 = m_1/(2\xi)$ and $w_2 = 1 - m_1/\xi$. The formula gives $\xi^- = \alpha_0 = m_1$, the one-point Gauss rule (node $m_1$, weight $1$), and $w_2 > 0$ if and only if $\xi > m_1$. The fiber is $\xi \in (m_1, T]$, and the free moment is $m_2 = m_1 \xi$, strictly increasing as Corollary \ref{cor:first_unmatched_odd} predicts.
\end{example}

\begin{remark}[Both parities]
    For every $K \geq 2$ the construction produces a one-parameter family sharing $R$ with $m_0, \dots, m_{K-2}$ matched and $m_{K-1}$ strictly increasing in $\xi$ (Corollaries \ref{cor:first_unmatched_even} and \ref{cor:first_unmatched_odd}). For $K$ even both endpoints are Gauss--Radau rules (nodes pinned at $0$ and at $T$). For $K$ odd the upper endpoint is Gauss--Radau at $T$, and the lower endpoint is the Gauss rule, where the configuration collapses to $K-1$ communities because the middle ray loses its mass.
\end{remark}

\subsection{Motif Densities Along the Fiber}
\label{sec:fiber_motifs}

Fix $K \geq 2$ and an admissible reference configuration $\tilde{\Theta}$ with radial law $R$. Write $\Theta_\xi$ for the configuration produced by Definition~\ref{def:construction_even} ($K$ even) or Definition~\ref{def:construction_odd} ($K$ odd), and
$$\Phi(\tilde{\Theta}) \coloneq \{ \Theta_\xi : \xi \in (\xi^-, \xi^+] \}$$
for the \emph{fiber} through $\tilde{\Theta}$, with $\xi^\pm$ from Theorem~\ref{thm:xi_admissibility} or Theorem~\ref{thm:xi_admissibility_odd}. Every member of $\Phi(\tilde{\Theta})$ is admissible, has $K$ communities (distinct slopes, positive weights), and has radial law $R$. By Corollaries~\ref{cor:uniqueness} and~\ref{cor:uniqueness_odd}, $\Phi(\tilde\Theta)$ is exactly the set of admissible $K$-community configurations with radial law $R$ whose slope moments $m_0, \dots, m_{K-2}$ agree with those of $\tilde{\Theta}$.

By Corollaries~\ref{cor:first_unmatched_even} and~\ref{cor:first_unmatched_odd}, for all $\xi_1 < \xi_2$ in $(\xi^-, \xi^+]$,
\begin{equation}
\label{eq:fiber_order}
    \Theta_{\xi_1} \sim_{K-2} \Theta_{\xi_2}, \qquad \Theta_{\xi_1} \sim^o_{K-1} \Theta_{\xi_2}, \qquad m_{K-1}(\Theta_{\xi_1}) < m_{K-1}(\Theta_{\xi_2}),
\end{equation}
and $m_{K-1}(\Theta_\xi) = \sigma_K\, \xi + c_K$ is affine in $\xi$ with slope
$$\sigma_K = |f_{k-1}|^2 \ (K = 2k), \qquad \sigma_K = m_1 |f'_{k-1}|^2 \ (K = 2k+1).$$
So $\mathrm{ord}(\Theta_{\xi_1}, \Theta_{\xi_2}) = K-2$ for every pair of distinct fiber points. Everything below follows from \eqref{eq:fiber_order} and Section~\ref{sec:moments_motifs}.

\subsubsection{Matched motifs}

\begin{theorem}[Matched motifs] \status{proved}
\label{thm:matched_motifs}
    Let $\Theta, \Theta' \in \Phi(\tilde{\Theta})$. Then $t(F;\Theta) = t(F;\Theta')$ for every graph $F$ with $h(F) \leq K-2$. In particular this holds for
    \begin{enumerate}
        \item every graph on at most $2K-2$ vertices, i.e.\ $\mathrm{Cen}_{2K-2}$ is constant on $\Phi(\tilde{\Theta})$;
        \item every graph of maximum degree at most $2K-3$, regardless of its number of vertices;
        \item every forest (these do not depend on the slopes at all).
    \end{enumerate}
    The same holds for induced (graphlet) densities of order at most $2K-2$.
\end{theorem}
\begin{proof}
    By \eqref{eq:fiber_order}, $\Theta \sim_{K-2} \Theta'$, and both share $R$. Theorem~\ref{thm:motif_transfer}(1) gives the main claim. Item 1 is Corollary~\ref{cor:census_transfer}(1) with $j = K-2$. Item 2 is Proposition~\ref{prop:slope_index_bounds}(1): $\Delta(F) \leq 2K-3$ gives $h(F) \leq K-2$. Item 3 is $h(F) = 0$. The induced statement follows from the M\"obius remark in Section~\ref{sec:hom_density}.
\end{proof}

\begin{remark}
    Item 2 says the obstruction is controlled by degree, not size. For instance, for $K = 4$ every cycle, every cubic graph and every graph of maximum degree $5$ has constant density along the fiber, however large.
\end{remark}

\begin{corollary}[No fixed census order suffices] \status{proved}
\label{cor:diagonal}
    For every $m \geq 2$ there exist $K$ and a fiber $\Phi(\tilde{\Theta})$ containing more than one configuration (indeed a continuum) on which $\mathrm{Cen}_m$ is constant.
\end{corollary}
\begin{proof}
    Take $K$ with $2K - 2 \geq m$ and any admissible reference with $K$ communities (for example equal weights, slopes $\pm$ equally spaced in $(0, \sqrt{T})$, and $R$ uniform on $(0, C)$ with $C < 1$). Its fiber is a nondegenerate interval by Theorem~\ref{thm:xi_admissibility} or~\ref{thm:xi_admissibility_odd}, and distinct $\xi$ give distinct configurations (Proposition~\ref{prop:monotone_transport}). Apply Theorem~\ref{thm:matched_motifs}.
\end{proof}

\subsubsection{Smallest breaker motif}

\begin{theorem}[Breaker motifs] \status{proved}
\label{thm:breaker}
    Let $\xi_1 < \xi_2$ in $(\xi^-, \xi^+]$.
    \begin{enumerate}
        \item \emph{(Monotone separation.)} For every $F$ with $h(F) = K-1$, $t(F;\Theta_{\xi_1}) < t(F;\Theta_{\xi_2})$. So each such density is strictly increasing along the fiber, and in particular injective on it.
        \item \emph{(Minimal order.)} No graph on at most $2K-2$ vertices separates $\Theta_{\xi_1}$ from $\Theta_{\xi_2}$. A graph $F$ on exactly $2K-1$ vertices separates them if and only if $F$ has a vertex $v$ adjacent to all others such that every connected component of $F[N(v)]$ has an even number of vertices. Hence $\mathrm{Cen}_{2K-1}(\Theta_{\xi_1}) \neq \mathrm{Cen}_{2K-1}(\Theta_{\xi_2})$.
        \item \emph{(Smallest breaker.)} Among separating graphs on $2K-1$ vertices, the friendship graph $F_{K-1}$ is the unique one, up to isomorphism, with the fewest edges ($3(K-1)$).
        \item \emph{(Signal.)} For $K \geq 3$, $t(F_{K-1};\Theta_\xi)$ is affine in $\xi$:
        $$t(F_{K-1};\Theta_{\xi_2}) - t(F_{K-1};\Theta_{\xi_1}) = \rho_{2K-2}\, \rho_2^{2K-2}\, m_1^{2K-2}\, \sigma_K\, (\xi_2 - \xi_1).$$
    \end{enumerate}
\end{theorem}
\begin{proof}
    (1) By \eqref{eq:fiber_order}, apply Theorem~\ref{thm:motif_transfer}(2) with $j = K-1$.

    (2) The first claim is Theorem~\ref{thm:matched_motifs}(1). A graph on $2K-1$ vertices has $h(F) \leq K-1$ (Proposition~\ref{prop:slope_index_bounds}(2)). If $h(F) \leq K-2$ it does not separate (Theorem~\ref{thm:matched_motifs}); if $h(F) = K-1$ it does, by (1). Proposition~\ref{prop:minimal_order} with $j = K-1$ identifies the graphs with $h(F) = K-1$. $F_{K-1}$ is one of them (Example~\ref{ex:friendship}), so the census of order $2K-1$ differs.

    (3) Let $F$ separate, with $v$ as in (2), and write $j = K-1$. $F$ has $2j$ edges at $v$. $F[N(v)]$ has $2j$ vertices, and since its components have even order it has no isolated vertices, so its degrees sum to at least $2j$ and it has at least $j$ edges. Hence $|E(F)| \geq 3j$. Equality forces every vertex of $F[N(v)]$ to have degree exactly $1$, i.e.\ $F[N(v)]$ is a perfect matching, and then $F \cong F_j$.

    (4) By Example~\ref{ex:friendship}, $t(F_{K-1};\Theta) = \rho_{2K-2}\rho_2^{2K-2}\sum_{i=0}^{K-1}\binom{K-1}{i} m_1^{2i} m_i$. For $K \geq 3$, $m_1$ is among the matched moments, so every term with $i \leq K-2$ is constant on the fiber, and the $i = K-1$ term is $\rho_{2K-2}\rho_2^{2K-2} m_1^{2K-2} m_{K-1}$ with $m_{K-1} = \sigma_K \xi + c_K$.
\end{proof}

\begin{remark}[$K = 2$]
    For $K = 2$ the only matched moment is $m_0$, and $m_1 = \xi$ itself varies. Items (1)--(3) still hold: the edge density ($2$ vertices) is constant and the triangle $F_1$ is the unique separator on $3$ vertices. But $t(F_1;\Theta_\xi) = \rho_2^3(1 + \xi^3)$ is cubic rather than affine in $\xi$.
\end{remark}

\begin{remark}[Size of the separating signal] \status{proved}
    Every factor in item (4) is small. $m_1 = \mathbb{E}[s^2] \leq t_1 \leq T \leq 1$, and $\rho_d \leq C^d$ with $C^2 \leq 1/2$ whenever $T = 1$ (and $C^2 < 1$ in general). So the leading factor $\rho_{2K-2}\rho_2^{2K-2} m_1^{2K-2}$ decays geometrically in $K$. The density difference that the census must resolve is therefore both of high order ($2K-1$ vertices) and geometrically small, which is the source of the census-detection cost discussed in Section~\ref{sec:census}%
.
\end{remark}

\begin{remark}
    Theorem~\ref{thm:breaker} says nothing about graphs with $h(F) \geq K$. Their densities involve $m_K, m_{K+1}, \dots$, which also move along the fiber, but not necessarily monotonically. Whether every such $F$ separates every pair of fiber points is left open. \status{open} It is not needed: $F_{K-1}$ alone already separates every pair.
\end{remark}

\subsection{Additional proofs}\label{app:new}

\paragraph{Proof of \cref{prop:bound}.} \status{proved}
By \cref{thm:even_expansion}, each term is $\prod_v m_{\deg_S(v)/2}$. Suppose it
contains $m_j$, so some $v$ has $\deg_S(v)=2j$. Then $v$ has $2j$ distinct
$S$-neighbours $u$, each with $\deg_S(u)\ge1$ and even, so $\deg_S(u)\ge2$.
Since $s^2\in[0,T]\subseteq[0,1]$, we have $m_r=\E[s^{2r}]\le\E[s^2]=m_1$ for
$r\ge1$, and $m_r\le m_0=1$ for all $r$. Bound the factor of $v$ by $m_j$,
those of its $2j$ neighbours by $m_1$, and the rest by $1$. For $F_j$, the
union of all $j$ triangles gives exactly $m_jm_1^{2j}$ (\cref{ex:friendship}).

\paragraph{Proof of \cref{cor:census_char}.} \status{proved}
($\Leftarrow$) This is \cref{thm:matched_motifs}(1). ($\Rightarrow$) Suppose
$\Theta$ shares $R$, has $K$ communities, is admissible and has the same
census of order $2K-2$ as $\tilde\Theta$, but $\Theta\notin\Phi(\tilde\Theta)$.
By the characterization of $\Phi(\tilde\Theta)$ in \cref{sec:fiber_motifs},
$m_d(\Theta)\ne m_d(\tilde\Theta)$ for some $d\le K-2$. Let $j$ be the smallest
such $d$; $j\ge1$ since $m_0=1$. Then $\Theta\sim^o_j\tilde\Theta$ and their
$m_j$ differ. The friendship graph $F_j$ has $h(F_j)=j$ and $2j+1\le2K-3$
vertices, so \cref{thm:motif_transfer}(2) gives
$t(F_j;\Theta)\ne t(F_j;\tilde\Theta)$, a contradiction.

\paragraph{Proof of \cref{cor:mpnn}.} \status{proved}

(1) \emph{Trees.} A tree $T$ is acyclic, so its cycle space is trivial and
$\mathcal E(T)=\{\emptyset\}$. By \cref{thm:even_expansion}, $\tau(T;\pi)=1$, so
by \cref{lem:factorization}, $t(T;\Theta)=\prod_v\rho_{\deg_T v}$, which depends
on $R$ only.

(2) \emph{Refinement.} On the graphon with points $z=(c,x)$ and kernel
$W(z,z')=xx'(1+s_cs_{c'})$, the round-$0$ color is constant, and the
round-$(t+1)$ color of $z$ is the measure
$A\mapsto\int\mathbf 1\{\mathrm{col}_t(z')\in A\}\,W(z,z')\,dP(z')$
\citep{grebik2022fractional,boker2023fine}. Suppose
$\mathrm{col}_t(c,x)=f_t(x)$. The community of $z'$ is independent of its
radius and $\E[s]=0$, so this measure is
$x\,(1+s_c\E[s])\int\mathbf 1\{f_t(x')\in A\}\,x'\,dR(x')=x\int\mathbf
1\{f_t(x')\in A\}\,x'\,dR(x')$, which depends on $x$ only. By induction every
color is a function of the radius, so configurations sharing $R$ have the same
distribution of colors.

(3) \emph{Networks.} The model is a graphon on the standard probability space
$\Omega=[K]\times(0,C)$ with law $\pi_\Theta\otimes R$. Moving it to $[0,1]$ by a
measure-preserving map changes neither tree densities nor the distribution of
colors \citep{lovasz2012large}. \citet{boker2023fine} consider networks with
Lipschitz updates $h^{(t)}(x)=\varphi_t\bigl(\int W(x,y)\,h^{(t-1)}(y)\,dy\bigr)$,
the graphon form of sum aggregation normalized by the number of nodes,
readout $\psi\bigl(\int h^{(L)}\bigr)$, and no node features. For this class,
the output is a function of the distribution of colors, and two graphons get
equal outputs from every network iff their tree densities agree
\citep[Theorems~12 and~28]{boker2023fine}. Apply (1) and (2). The argument of
(2) also covers aggregation normalized by degree, since the degree $x\rho_1$
is a function of the color; that class is outside the cited theorem.

\section{Census Route: Details}\label{app:census}
\paragraph{Homomorphism counts are unbiased for the separation.} For a graph
$H$ on $m$ vertices, grouping maps by which vertices coincide gives
$\E[\hom(H,G)/n^m]=\sum_{P}\frac{(n)_{|P|}}{n^m}\,t(H/P)$ over partitions $P$ of
$V(H)$. Every proper quotient $H/P$ has at most $m-1$ vertices; for $H=F_{K-1}$
($m=2K-1$) its density is therefore constant along the fiber
(\cref{thm:matched_motifs}). In the difference between two fiber
configurations every degenerate term cancels, leaving
$\frac{(n)_m}{n^m}\,(t_X(H)-t_Y(H))$, and the known factor
$(n)_m/n^m=1-O(m^2/n)$ is divided out. The friendship ladder then reduces to
per-vertex triangle counts, $\hom(F_r,G)=\sum_v((A^3)_{vv})^r$. Counts are
accumulated in integer arithmetic, since the $K=4$ separation is below float32
resolution.

\paragraph{Estimators P2, P2$'$, P2$''$.} P2 takes $m_1$ from
$\lambda_2/\lambda_1$, inverts $t(F_2)$ and $t(F_3)$ for $m_2,m_3$, and
applies Prony. P2$'$ estimates $m_2$ instead from $V=t(F_2)-t(C_3)^2$, the
centred second moment of per-vertex triangle counts, in which the fluctuation
of the mean triangle count shared by the two densities cancels; the radial
moments are estimated from star densities on the same graph, and each density
is divided by its exact finite-$n$ expectation. P2$''$ minimizes
$\sum_{r=1}^{3}(\hat t(F_r)-t(F_r;\theta))^2/\mathrm{SD}_r^2$ over the
admissible set, with $\mathrm{SD}_r$ the replicate standard deviation of the
normalized density.
The $m_2$-dependent part of $t(F_2)/\text{base}$ is $0.013\%$
of the total against about $3\%$ Monte Carlo noise, predicting
$\mathrm{SD}(\hat m_2)=13.7$ against a measured $12.6$, $234\times$ the true
$m_2$. Of $12{,}600$ replicates at the published configuration, $49.9\%$
failed Hankel positivity and the rest returned atoms or weights out of range.
A sweep of $m_1$ from $0.217$ to $0.85$ never exceeds a $2\%$ valid rate for P2
or P2$'$; at $m_1=0.85$, $n=32{,}000$, P2$'$ has $\mathrm{SNR}(m_2)=126.9$ and
$\mathrm{SNR}(m_3)=1.61$ and still returns $0$ of $200$.

\begin{figure}[H]\centering
\includegraphics[width=0.95\textwidth]{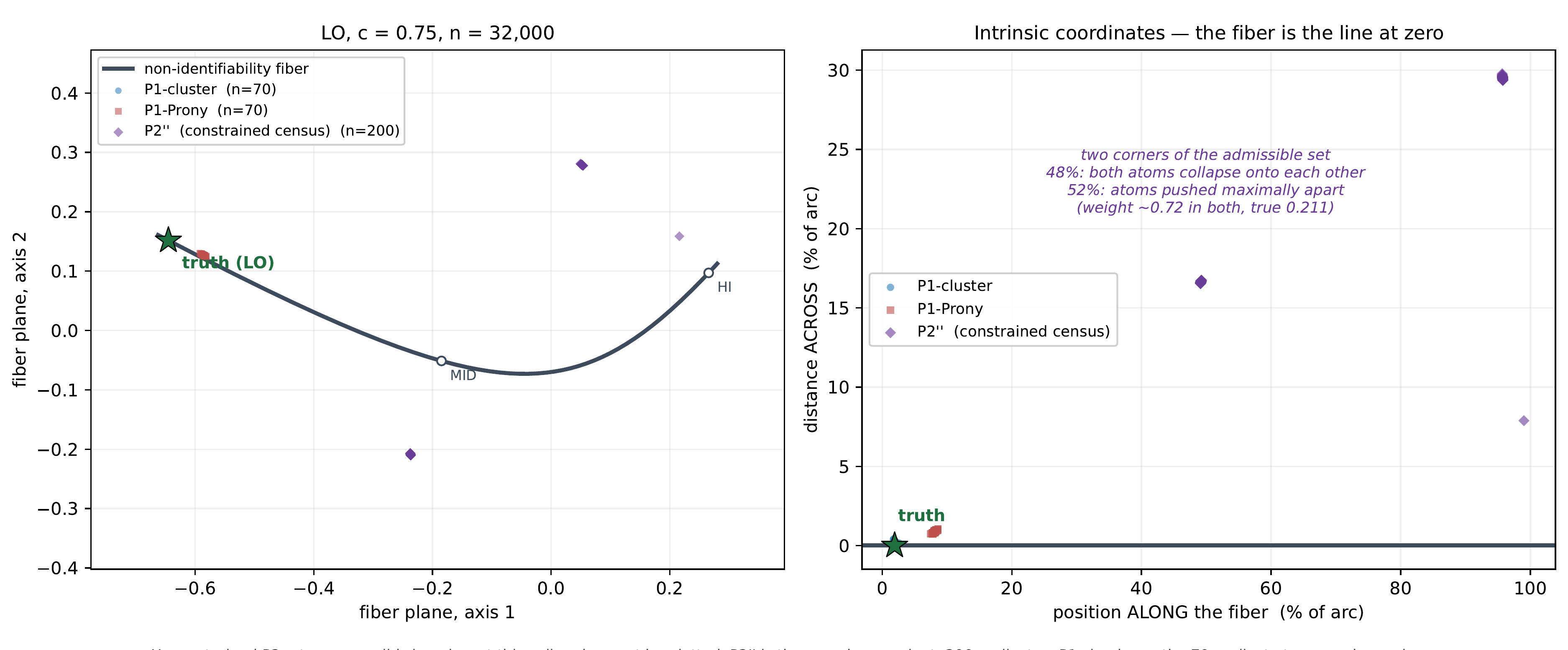}%
\caption{Three estimators on identical graphs at LO, $C=0.75$,
$n=32{,}000$. Unconstrained P2 returns no valid slope law here and is not
shown. P2$''$ fits sit on the boundary of the admissible set; its $\pm23\%$
replicate spread is the gap between two point masses, not dispersion.}\label{fig:three}
\end{figure}

\paragraph{P2$''$ is valid, bimodal, and on the boundary.}
At LO, $C=0.75$, $n=32{,}000$ ($200$ replicates) the fits fall on two point
masses (\cref{tab:p2pp}, \cref{fig:three}); the objective, $\chi^2$-like on
its three residuals with expectation about $3$, is about $1.2$ for the truth and
for both corners, and all refinements converge. The reported $\pm23\%$ spread
is the gap between point masses, not dispersion.

\begin{table}[H]\centering\small
\caption{P2$''$ fits at LO, $C=0.75$, $n=32{,}000$.}\label{tab:p2pp}
\begin{tabular}{lccccc}\toprule
fit & share & $\tau_1$ & $\tau_2$ & $\alpha_1$ & objective\\\midrule
atoms collapsed & $48.0\%$ & 0.218 & 0.218 & 0.720 & 1.197\\
atoms maximally apart & $51.5\%$ & 0.0000 & 0.7777 & 0.720 & 1.240\\
truth & --- & 0.0115 & 0.2725 & 0.2111 & 1.232\\\bottomrule
\end{tabular}
\end{table}

\section{Held-Out Scores}\label{app:scores}
The held-out oracle score holds out $5\%$ of node pairs and takes the true
latent positions. For each candidate it snaps every node to the candidate's
nearest ray (by $|\text{slope}|$), projects the position onto that ray, and
scores the held-out pairs by the log-loss of the resulting inner products,
clipped to $[10^{-6},1-10^{-6}]$; the mask, outcomes and positions are shared,
and only the candidate's slopes differ. For the generating configuration the
snap is the identity, so its predictions are the true edge probabilities and
it must win in expectation (Gibbs' inequality). The analytic
benchmark is the population expectation of the per-pair log-loss gap,
computed by Gauss--Legendre quadrature over the radial law and exact summation
over slopes; it involves no graph, and $n$ enters only through the conversion
to a $t$-statistic. The oracle's margin matches it at ratios $0.93$--$1.04$
across all $36$ cells.

The reconstruction (``plug-in'') score uses the same mask, sets the held-out
entries of $A$ to zero without rescaling (which shrinks the embedding by a
factor of about $\sqrt{0.95}$, equally for both candidates), embeds, and then
applies the same snap to the estimated positions: it snaps each node to the candidate's nearest ray, projects to a
radius, and scores the held-out pairs by the log-loss of the resulting inner
products. Both candidates share
one embedding and one mask and differ only in the snap; it compares two
constrained reconstructions of one estimated point cloud and is not a
likelihood ratio, so Gibbs' inequality does not apply. It is correct at
$C=0.75$ and inverts below, always toward the same arm; the candidate's weights
never enter it. The oracle decomposition (analytic benchmark $\to$ oracle
$\to$ plug-in, each adding one error source) reclassifies the $23$ cells in
which the plug-in score preferred the wrong configuration as construction
artifacts.

\section{Spectral Pipeline: Details}\label{app:spectral}
\begin{table}[H]
\caption{P1-cluster at $C=0.3$, $n=32{,}000$: intrinsic signed along-fiber
displacement (bias) and spread, \% of arc ($+$ toward the pair-B end).}\label{tab:bias}
\centering\small
\begin{tabular}{lcccc}
\toprule
config & $w_{\min}$ & displacement & spread & ratio\\
\midrule
LO & 0.2111 & $+2.65$ & $0.232$ & 11\\
MID & 0.3140 & $-4.96$ & $0.341$ & 15\\
HI & 0.0375 & $-31.4$ & $4.67$ & 6.7\\
pair B & 0.0361 & $-33.1$ & $4.42$ & 7.5\\
\bottomrule
\end{tabular}
\end{table}

\begin{figure}[H]\centering
\includegraphics[width=0.8\textwidth]{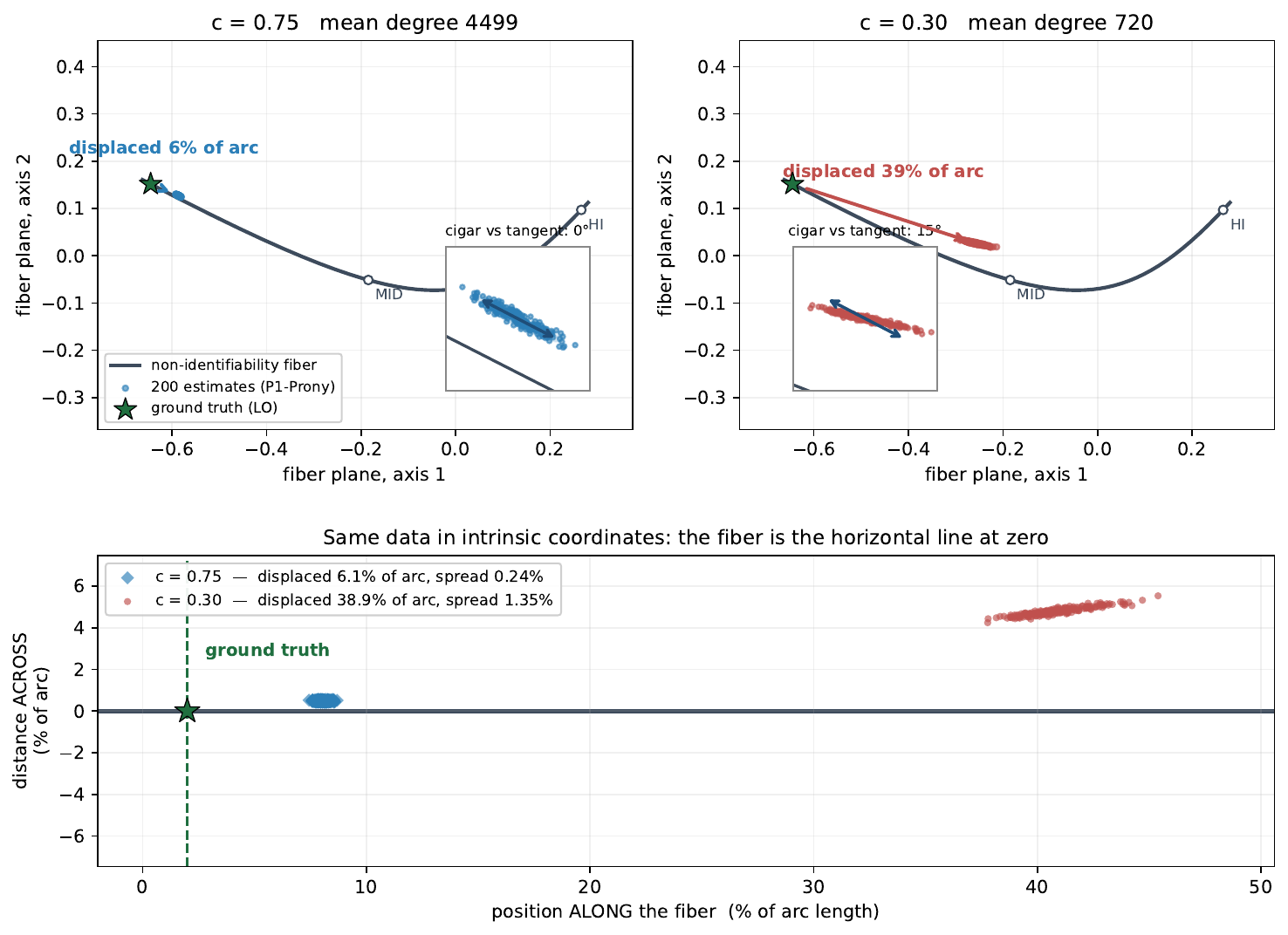}
\caption{P1-Prony at LO, $n=32{,}000$: displaced $6.1\%$ at $C=0.75$ and
$38.9\%$ at $C=0.3$ (intrinsic mean displacement, $200$ replicates per
cell).}\label{fig:prony}
\end{figure}

\begin{table}[H]\centering\small
\caption{The two tilts.}\label{tab:tilts}
\begin{tabular}{lll}\toprule
 & P1-cluster & P1-Prony\\\midrule
tilt & heavy tail of $\hat T$ crosses 2-means boundary & third-moment inflation ($+40$--$64\%$)\\
error type & classification & estimation of a moment\\
worst at & small minority weight (steep ray) & wherever the arc has room above\\
sign & toward $\alpha_1=0.5$ fold (90/90, 162/186) & toward larger $m_3$ (229/276)\\
\end{tabular}
\end{table}

\paragraph{Tube around the fitted ambiguity set.} A tube with radius
propagated from the replicate covariance of $(\hat m_1,\hat m_2)$ through the
level-set Jacobian covers at least $80\%$ while staying informative in $1$ of
$16$ groups at $z=3$ and $0$ of $16$ at $z=1,2$; the one success is the cell
where the point estimate is already $99.9\%$ accurate, with bias/spread
$1.47$. It fails by not covering, not by becoming vacuous.

\paragraph{Bias scaling.} Across-fiber distance at $C=0.75$:
$4.67\%\to0.865\%\to0.161\%$ of arc over $16\times$ in $n$ (slope $-1.21$);
bias/spread $6.62\to3.85\to1.47$. At $C=0.3$ the clean scaling degrades.

\section{Published Pipelines and the Factorial}\label{app:pipelines}
All pipelines use the same $d=2$ ASE as P1 and cluster every row (the
published recipes have no radius filter). SCORE takes $\hat\eta_2/\hat\eta_1$
with $\hat\eta_1$ given the Perron sign, truncates at $\pm\log n$, and runs
one-dimensional $k$-means with $K=4$ and $10$ restarts
\citep{jin2015fast}. Row-normalized spectral clustering runs $k$-means on unit
rows \citep{qin2013regularized,lei2015consistency}. Spherical coordinates use
the authors' transform with a Gaussian mixture, $K=4$, $10$ initializations
\citep{sannapassino2022spectral}; isolated nodes give zero rows, on which the
transform is undefined, so they are fitted out and assigned afterward. The
shared plug-in sets each cluster's slope to $\sum x_1x_2/\sum x_1^2$, pairs the
clusters by $|$slope$|$, and takes $\tau$ as the mean squared slope per pair and
$\alpha_1$ as the shallow pair's share. Label accuracy (best permutation) at HI,
$C=0.3$: SCORE $95.3\%$, row-normalized $75.4\%$, spherical $88.4\%$.
SCORE's failure at LO is bimodal: at $C=0.3$, $6.5\%$ of replicates land about
$81\%$ of the arc away; at $C=0.5$, $34.9\%$ land about $105\%$ away, and $10\%$
of fits leave the admissible set. Spherical coordinates with a GMM is also
bimodal at LO in the dense control ($C=0.75$): $37\%$ of replicates land about
$38\%$ of the arc away.

\begin{table}[H]\centering\small\setlength{\tabcolsep}{4pt}
\caption{Factorial of summary pipelines (shared slope plug-in, $R=200$): median
total error $|\hat\theta-\theta_0|$, \% of arc, by cell (mean degree), and
fitted-to-true minority weight at HI (720). ``Filter'' is P1's small-radius
filter. The untruncated ratio matches the truncated one to within $0.3$ points
and is omitted.}\label{tab:factorial}
\begin{adjustbox}{max width=\textwidth}\begin{tabular}{lllrrrrrrrr}\toprule
feature & clusterer & filter & HI\,(720) & pair B\,(720) & HI\,(500) & $w_{\min}{=}0.03$\,(500) & LO\,(720) & LO\,(500) & MID\,(720) & $\hat w/w$\\\midrule
ratio, truncated & k-means & none & 4.2 & 4.07 & 7.74 & 7.42 & 3.09 & 47.9 & 3.67 & 1.97\\
ratio, truncated & k-means & P1 & 26.7 & 29.1 & 43.2 & 52.3 & 2.16 & 3.73 & 5.3 & 3.91\\
ratio, truncated & GMM & none & 10.1 & 10.3 & 12.3 & 14.6 & 4.34 & 4.61 & 29.1 & 2.72\\
ratio, truncated & GMM & P1 & 13.2 & 14.2 & 21.2 & 25 & 8.05 & 8.88 & 0.956 & 2.53\\
\addlinespace[2pt]
angle & k-means & none & 40.6 & 42.4 & 50.3 & 58.1 & 2.12 & 3.28 & 7.82 & 6.89\\
angle & k-means & P1 & 54.9 & 55.8 & 58.2 & 65.4 & 1.48 & 2.54 & 8.51 & 10.6\\
angle & GMM & none & 26.8 & 27.7 & 32.2 & 37.6 & 0.866 & 5.15 & 2.48 & 4.02\\
angle & GMM & P1 & 29.1 & 30.4 & 39.8 & 46.6 & 1.07 & 1.38 & 34.5 & 3.66\\
\addlinespace[2pt]
unit vector & k-means & none & 44.1 & 44.7 & 50.5 & 57.9 & 2.16 & 3.32 & 7.98 & 7.49\\
unit vector & k-means & P1 & 54.6 & 56.1 & 58.1 & 65.5 & 1.48 & 2.55 & 8.56 & 10.5\\
unit vector & GMM & none & 41.4 & 42.7 & 62.5 & 79.6 & 7.8 & 7.71 & 32.6 & 5.38\\
unit vector & GMM & P1 & 60.5 & 61.1 & 62 & 66.2 & 2.1 & 3.04 & 8.9 & 11.8\\
\bottomrule\end{tabular}\end{adjustbox}\end{table}

\section{Model-Based Fits}\label{app:rayem}
\paragraph{RayEM.} Each embedded row is modeled as
$\hat x_i=r_i(1,s_{c_i})+\varepsilon_i$ with
$\varepsilon_i\sim N(0,\hat\Sigma(\hat x_i)/n)$, where
$\Sigma(x)=\Delta^{-1}\E[(x^\top Y-(x^\top Y)^2)YY^\top]\Delta^{-1}$ is the ASE
central-limit covariance \citep{athreya2016limit}, $Y$ is an independent latent
position and $\Delta=\E[YY^\top]$. Replacing expectations over $Y$ by averages
over the rows, $\hat\Delta=n^{-1}\sum_j\hat x_j\hat x_j^\top$ and
$\E[(x^\top Y)^q\,YY^\top]\approx n^{-1}\sum_j(x^\top\hat x_j)^q\,\hat x_j\hat x_j^\top$
for $q=1,2$, which are contractions of the rows' third- and fourth-moment
tensors, so all of $\hat\Sigma$ costs $O(n)$; no ground truth, radial law or $C$ is used. Slopes are $\pm s_a$,
$\pm s_b$ with reflected pairs sharing a weight. Each row's radius is profiled
out per ray in the Mahalanobis metric. The E-step assigns responsibilities
from the profiled residuals and the weights; the M-step updates the weights
and then $s_a,s_b$, which separate into two bounded one-dimensional problems.
Seven starts are used (P1-cluster, P1-Prony, five random), and the best
objective is kept. Rows are filtered as for P1, because $\hat\Sigma$ vanishes
near the origin; rows whose plug-in matrix is not positive definite get a small
ridge. \emph{Validation.} Under its own model (synthetic Gaussian rows at
$n=32{,}000$) RayEM is unbiased to $1\%$ of the arc. We pre-specified that it
recover $\theta$ to $10^{-6}$ from noise-free rows. It does not: errors are
$2$--$3\times10^{-3}$, because the plug-in noise level is nonzero even for
noise-free rows, so the profile objective is not stationary at the truth. The
criterion holds in the zero-noise limit; we changed it to that after the
first run and report both. The plug-in
covariance matches the realized embedding error to $2\%$ (ratio $0.98$).

\paragraph{The authors' Bayesian LSBM.} We ran the Gibbs sampler of
\citet{sannapassino2022latent} from the authors' public code, with curves
restricted to lines through the origin, on $20$ graphs at each of HI
($C=0.3$, $n=32{,}000$) and $w_{\min}=0.03$ ($C=0.5$, $n=8000$). Chains ran until the
split-half drift of every component of $\hat\theta$ fell below $10^{-3}$, with a
$3$-hour cap; $17$ of $20$ and $19$ of $20$ chains converged. Over all $20$
chains, converged or not, along-fiber displacement was $-0.33\%$ and $-1.76\%$ (RMSE $0.81\%$, $2.32\%$), against
$-31.3\%$ and $-54.5\%$ for P1-cluster on the same graphs.

\section{Negative Results and Pre-Registered Predictions}\label{app:negative}
\paragraph{Richardson correction.} Re-estimating on nested node subsamples
($n/2$, $n/4$; four draws each) and extrapolating as $1/n\to0$ does not remove
the displacement. At the moment level, Prony on corrected moments is unstable:
no valid slope law at LO ($C=0.3$), and HI moves from $-9.6\%$ to $-22.5\%$. At
the estimate level, P1-cluster's displacement at HI falls from $-31.4\%$ to
$-16.8\%$, but in the dense control cells the correction overshoots where the
bias is already small: the bias there shrinks faster than $1/n$. The overshoot
is systematic (at HI, $C=0.75$, $+1.36\%$ against a baseline of $-0.24\%$, many
standard errors), not noise from using four subsample draws.

\paragraph{A single-graph detector.} The drift of the estimate between $n/2$
and $n$ separates displaced from accurate cells with AUC $0.96$, but its sign
agrees with the displacement in only $67\%$ of replicates ($50\%$ for
P1-Prony), and at a threshold calibrated on the dense cells it flags every
replicate of P1-cluster's safe low-density cells. It detects low density, not
displacement, and we do not recommend it.

\paragraph{Predictions.} Before each stage we recorded predictions with
pass rules. Held: estimate-level correction only partly removes P1-cluster's
displacement; RayEM removes it, with its residual error axis within
$30^\circ$ of the tangent ($8.3^\circ$); RayEM with correction is within
$5\%$ at HI and LO. Failed: moment-level correction fixes Prony (no valid
output at LO); the detector meets AUC $\ge0.9$ \emph{and} $90\%$ sign agreement;
SCORE and row-normalized clustering both slide by more than $20\%$ at HI
(SCORE does not); untruncated ratios with $k$-means slide by more than $20\%$
(they do not); the small-radius filter halves the slide of angle and
unit-vector features (it does not). The prediction texts were fixed in code
and time-stamped before the runs they concern.

\section{Per-Cell Tables and Provenance}\label{app:tables}
Every localization number is intrinsic: position and displacement are arc
length along the fiber, in \% of the reference arc ($1.0571$ in $\theta$, the
admissible arc at $C=0.75$), and every displacement is reported with its
spread and RMSE. Configurations with $s/\mathrm{arc}>1$ (e.g.\ $w_{\min}=0.03$)
are admissible only for $C\le1/\sqrt2$, where the admissible fiber is longer.
\Cref{tab:safe,tab:ambiguity,tab:tube,tab:keycells} condense the per-cell
results; the full tables for all $552$ (estimator, configuration, $C$, $n$)
cells of each experiment are provided as ancillary files with this article, in PDF and CSV. In these tables, P1-c is P1-cluster and P1-P is P1-Prony.

{\scriptsize\setlength{\tabcolsep}{3pt}
\begin{longtable}{llrrrrrrrrrr}
\caption{Largest along-fiber displacement and safe region, per estimator, branch and $(C,n)$. Displacement is intrinsic and signed (\% of the reference arc; $+$ toward the pair-B end), with the relative change in $m_3$ at the landing point. \emph{Safe} means $|$displacement$|<5\%$ for this and every larger minority weight on the same branch, stated on the true and on the fitted weight scale; ``none means no safe weight on that branch. \emph{unsafe of}: unsafe points out of points on the branch. Mean degree $45$ is estimator breakdown (off-curve error) and is not comparable to the other rows.}\label{tab:safe}\\
\toprule est. & minority ray & $C$ & $n$ & deg. & largest disp. & $\Delta m_3$ & at $w_{\min}$ & safe from $w_{\min}\ge$ & safe from $\hat w_{\min}\ge$ & unsafe & of\\\midrule\endfirsthead
\toprule est. & minority ray & $C$ & $n$ & deg. & largest disp. & $\Delta m_3$ & at $w_{\min}$ & safe from $w_{\min}\ge$ & safe from $\hat w_{\min}\ge$ & unsafe & of\\\midrule\endhead\bottomrule\endfoot
P1-c & shallow & 0.75 & 2000 & 281 & +9.23 & +3.32 & 0.2732 & 0.4458 & 0.4796 & 8 & 10\\
\addlinespace[2pt]
P1-c & shallow & 0.75 & 8000 & 1125 & +1.52 & +0.529 & 0.3154 & 0.1936 & 0.1991 & 0 & 10\\
\addlinespace[2pt]
P1-c & shallow & 0.75 & 32000 & 4500 & +0.194 & +0.0684 & 0.3154 & 0.1936 & 0.1932 & 0 & 10\\
\addlinespace[2pt]
P1-c & shallow & 0.5 & 2000 & 125 & +29.1 & +11.1 & 0.1936 & none & -- & 10 & 10\\
\addlinespace[2pt]
P1-c & shallow & 0.5 & 8000 & 500 & +5.21 & +1.95 & 0.2732 & 0.3584 & 0.4009 & 3 & 10\\
\addlinespace[2pt]
P1-c & shallow & 0.5 & 32000 & 2000 & +0.718 & +0.252 & 0.3154 & 0.1936 & 0.1952 & 0 & 10\\
\addlinespace[2pt]
P1-c & shallow & 0.3 & 2000 & 45 & +70 & +27.9 & 0.1936 & none & -- & 10 & 10\\
\addlinespace[2pt]
P1-c & shallow & 0.3 & 8000 & 180 & +22.1 & +8.93 & 0.1936 & 0.4897 & 0.4812 & 9 & 10\\
\addlinespace[2pt]
P1-c & shallow & 0.3 & 32000 & 720 & +3.39 & +1.16 & 0.3154 & 0.1936 & 0.2109 & 0 & 10\\
\addlinespace[2pt]
P1-c & steep & 0.75 & 2000 & 281 & -50.3 & -21.6 & 0.03613 & 0.4 & 0.4472 & 16 & 20\\
\addlinespace[2pt]
P1-c & steep & 0.75 & 8000 & 1125 & -7.74 & -4.04 & 0.03613 & 0.07 & 0.09453 & 4 & 20\\
\addlinespace[2pt]
P1-c & steep & 0.75 & 32000 & 4500 & -0.365 & -0.158 & 0.19 & 0.03613 & 0.03641 & 0 & 20\\
\addlinespace[2pt]
P1-c & steep & 0.5 & 2000 & 125 & -62 & -25.1 & 0.03 & none & -- & 18 & 21\\
\addlinespace[2pt]
P1-c & steep & 0.5 & 8000 & 500 & -53.5 & -23.2 & 0.03 & 0.36 & 0.4159 & 16 & 21\\
\addlinespace[2pt]
P1-c & steep & 0.5 & 32000 & 2000 & -2.73 & -1.38 & 0.03 & 0.03 & 0.03302 & 0 & 21\\
\addlinespace[2pt]
P1-c & steep & 0.3 & 2000 & 45 & +39.5 & +15.1 & 0.495 & none & -- & 18 & 21\\
\addlinespace[2pt]
P1-c & steep & 0.3 & 8000 & 180 & -63.8 & -25.5 & 0.03 & 0.4 & 0.4603 & 17 & 21\\
\addlinespace[2pt]
P1-c & steep & 0.3 & 32000 & 720 & -45.5 & -21 & 0.03 & 0.314 & 0.3692 & 14 & 21\\
\midrule
P1-P & shallow & 0.75 & 2000 & 281 & +76 & +32.1 & 0.1936 & none & -- & 10 & 10\\
\addlinespace[2pt]
P1-P & shallow & 0.75 & 8000 & 1125 & +25.2 & +9.42 & 0.2111 & none & -- & 10 & 10\\
\addlinespace[2pt]
P1-P & shallow & 0.75 & 32000 & 4500 & +6.17 & +2.28 & 0.2732 & 0.4458 & 0.4689 & 8 & 10\\
\addlinespace[2pt]
P1-P & shallow & 0.5 & 2000 & 125 & +116 & +62.9 & 0.1936 & none & -- & 10 & 10\\
\addlinespace[2pt]
P1-P & shallow & 0.5 & 8000 & 500 & +50.9 & +18.3 & 0.1936 & none & -- & 10 & 10\\
\addlinespace[2pt]
P1-P & shallow & 0.5 & 32000 & 2000 & +14.9 & +5.7 & 0.2323 & none & -- & 10 & 10\\
\addlinespace[2pt]
P1-P & shallow & 0.3 & 2000 & 45 & +121 & +67 & 0.1936 & none & -- & 10 & 10\\
\addlinespace[2pt]
P1-P & shallow & 0.3 & 8000 & 180 & +99.7 & +50.4 & 0.1936 & none & -- & 10 & 10\\
\addlinespace[2pt]
P1-P & shallow & 0.3 & 32000 & 720 & +39.4 & +14.4 & 0.1936 & none & -- & 10 & 10\\
\addlinespace[2pt]
P1-P & steep & 0.75 & 2000 & 281 & +51 & +22.6 & 0.495 & none & -- & 17 & 20\\
\addlinespace[2pt]
P1-P & steep & 0.75 & 8000 & 1125 & +14.9 & +4.31 & 0.495 & none & -- & 12 & 20\\
\addlinespace[2pt]
P1-P & steep & 0.75 & 32000 & 4500 & +3.97 & +1.1 & 0.495 & 0.03613 & 0.04529 & 0 & 20\\
\addlinespace[2pt]
P1-P & steep & 0.5 & 2000 & 125 & +84.1 & +45.6 & 0.495 & none & -- & 21 & 21\\
\addlinespace[2pt]
P1-P & steep & 0.5 & 8000 & 500 & +30.4 & +10.2 & 0.495 & none & -- & 17 & 21\\
\addlinespace[2pt]
P1-P & steep & 0.5 & 32000 & 2000 & +9.07 & +2.55 & 0.495 & none & -- & 8 & 21\\
\addlinespace[2pt]
P1-P & steep & 0.3 & 2000 & 45 & +89.1 & +49.1 & 0.495 & none & -- & 21 & 21\\
\addlinespace[2pt]
P1-P & steep & 0.3 & 8000 & 180 & +70.7 & +36.3 & 0.495 & none & -- & 21 & 21\\
\addlinespace[2pt]
P1-P & steep & 0.3 & 32000 & 720 & +22.3 & +6.84 & 0.495 & none & -- & 15 & 21\\
\end{longtable}}

\begin{table}[H]\centering\scriptsize\setlength{\tabcolsep}{3pt}
\caption{Coverage of the fitted ambiguity set (the fiber through the fit's own $(\hat m_1,\hat m_2)$) at a tolerance of $1\%$ of the arc, against the point estimate and a $\chi^2_3$ 95\% ellipsoid built from the true replicate covariance. Coverage is measured, not guaranteed.}\label{tab:ambiguity}
\begin{adjustbox}{max width=\textwidth}{\begin{tabular}{lrrrrrlrrr}\toprule est. & $C$ & $n$ & deg. & cells & set @1\% & tol.\ for $\ge$80\% & cells $<$80\% & point @1\% & 95\% ellipsoid\\\midrule
P1-c & 0.75 & 2000 & 281 & 30 & 0.2 & $>$\,5 (none) & 30 & 0.0 & 0.0\\
\addlinespace[2pt]
P1-c & 0.75 & 8000 & 1125 & 30 & 68.0 & 2 of arc & 16 & 8.6 & 0.0\\
\addlinespace[2pt]
P1-c & 0.75 & 32000 & 4500 & 30 & 100.0 & 0.25 of arc & 0 & 99.9 & 5.1\\
\addlinespace[2pt]
P1-c & 0.5 & 2000 & 125 & 31 & 0.0 & $>$\,5 (none) & 31 & 0.0 & 0.0\\
\addlinespace[2pt]
P1-c & 0.5 & 8000 & 500 & 31 & 6.2 & 5 of arc & 31 & 0.0 & 0.0\\
\addlinespace[2pt]
P1-c & 0.5 & 32000 & 2000 & 31 & 100.0 & 1 of arc & 0 & 45.8 & 0.0\\
\addlinespace[2pt]
P1-c & 0.3 & 2000 & 45 & 31 & 0.0 & $>$\,5 (none) & 31 & 0.0 & 0.0\\
\addlinespace[2pt]
P1-c & 0.3 & 8000 & 180 & 31 & 0.0 & $>$\,5 (none) & 31 & 0.0 & 0.0\\
\addlinespace[2pt]
P1-c & 0.3 & 32000 & 720 & 31 & 22.8 & 5 of arc & 25 & 0.0 & 0.0\\
\midrule
P1-P & 0.75 & 2000 & 281 & 30 & 0.0 & $>$\,5 (none) & 30 & 0.0 & 0.0\\
\addlinespace[2pt]
P1-P & 0.75 & 8000 & 1125 & 30 & 0.0 & 5 of arc & 30 & 0.0 & 0.0\\
\addlinespace[2pt]
P1-P & 0.75 & 32000 & 4500 & 30 & 83.0 & 1 of arc & 8 & 3.3 & 0.0\\
\addlinespace[2pt]
P1-P & 0.5 & 2000 & 125 & 31 & 0.0 & $>$\,5 (none) & 31 & 0.0 & 0.0\\
\addlinespace[2pt]
P1-P & 0.5 & 8000 & 500 & 31 & 0.0 & $>$\,5 (none) & 31 & 0.0 & 0.0\\
\addlinespace[2pt]
P1-P & 0.5 & 32000 & 2000 & 31 & 8.2 & 5 of arc & 30 & 0.0 & 0.0\\
\addlinespace[2pt]
P1-P & 0.3 & 2000 & 45 & 31 & 0.0 & $>$\,5 (none) & 31 & 0.0 & 0.0\\
\addlinespace[2pt]
P1-P & 0.3 & 8000 & 180 & 31 & 0.0 & $>$\,5 (none) & 31 & 0.0 & 0.0\\
\addlinespace[2pt]
P1-P & 0.3 & 32000 & 720 & 31 & 0.0 & $>$\,5 (none) & 31 & 0.0 & 0.0\\
\bottomrule\end{tabular}}\end{adjustbox}\end{table}

\begin{table}[H]\centering\scriptsize\setlength{\tabcolsep}{3pt}
\caption{Thickened ambiguity set (a tube whose radius is propagated from the replicate covariance of $(\hat m_1,\hat m_2)$; it uses replicate information a practitioner does not have). $r_1$: tube radius at $z=1$; the tube at $z$ has radius $z\,r_1$, so $z$ is the multiplier on the propagated standard deviation. Coverage and size are reported together. \emph{adm.\ length}: admissible fiber length at that $C$, \% of the reference arc ($>100\%$ for $C<0.75$ by construction).}\label{tab:tube}
\begin{adjustbox}{max width=\textwidth}{\begin{tabular}{lrrlllrrll}\toprule est. & $C$ & $n$ & cover $z{=}1/2/3$ & $r_1$ (\% arc) & adm.\ length & truth$\to$spine & truth$\to$point & cells $<$80\% & vacuous\\\midrule
P1-c & 0.75 & 2000 & 0.0 / 0.8 / 6.2 & 0.783 (0.575--1.29) & 88 & 4.67 & 9.74 & 30 / 30 / 30 & 0 / 0 / 0\\
\addlinespace[2pt]
P1-c & 0.75 & 8000 & 1.1 / 13.4 / 37.1 & 0.225 (0.168--0.388) & 98.2 & 0.865 & 2.11 & 30 / 30 / 25 & 0 / 0 / 0\\
\addlinespace[2pt]
P1-c & 0.75 & 32000 & 23.8 / 70.9 / 93.7 & 0.102 (0.0759--0.151) & 100 & 0.161 & 0.283 & 30 / 20 / 1 & 0 / 0 / 0\\
\addlinespace[2pt]
P1-c & 0.5 & 2000 & 0.0 / 0.0 / 0.0 & 1.71 (0.767--2.54) & 92.6 & 11.1 & 23.4 & 31 / 31 / 31 & 0 / 0 / 0\\
\addlinespace[2pt]
P1-c & 0.5 & 8000 & 0.0 / 0.2 / 2.3 & 0.263 (0.191--0.522) & 115 & 2.43 & 7 & 31 / 31 / 31 & 0 / 0 / 0\\
\addlinespace[2pt]
P1-c & 0.5 & 32000 & 0.3 / 9.4 / 38.0 & 0.107 (0.0835--0.171) & 121 & 0.391 & 1.21 & 31 / 31 / 24 & 0 / 0 / 0\\
\addlinespace[2pt]
P1-c & 0.3 & 2000 & 0.0 / 0.0 / 0.0 & 6.41 (2.64--8.25) & 57 & 28.3 & 34 & 31 / 31 / 31 & 0 / 0 / 12\\
\addlinespace[2pt]
P1-c & 0.3 & 8000 & 0.0 / 0.0 / 0.0 & 0.579 (0.261--0.943) & 100 & 8.01 & 20.7 & 31 / 31 / 31 & 0 / 0 / 0\\
\addlinespace[2pt]
P1-c & 0.3 & 32000 & 0.0 / 0.0 / 1.3 & 0.134 (0.0917--0.242) & 117 & 1.58 & 5.16 & 31 / 31 / 31 & 0 / 0 / 0\\
\midrule
P1-P & 0.75 & 2000 & 0.0 / 0.0 / 0.0 & 1.86 (0.941--2.41) & 68.1 & 10.9 & 32.5 & 30 / 30 / 30 & 0 / 0 / 0\\
\addlinespace[2pt]
P1-P & 0.75 & 8000 & 0.0 / 0.0 / 0.0 & 0.269 (0.187--0.435) & 90.1 & 3.24 & 8.71 & 30 / 30 / 30 & 0 / 0 / 0\\
\addlinespace[2pt]
P1-P & 0.75 & 32000 & 0.0 / 0.1 / 1.3 & 0.102 (0.0759--0.152) & 97.3 & 0.848 & 3.14 & 30 / 30 / 30 & 0 / 0 / 0\\
\addlinespace[2pt]
P1-P & 0.5 & 2000 & 0.0 / 0.0 / 0.0 & 4.91 (2.06--6.23) & 64 & 21.5 & 63.5 & 31 / 31 / 31 & 0 / 0 / 0\\
\addlinespace[2pt]
P1-P & 0.5 & 8000 & 0.0 / 0.0 / 0.0 & 0.496 (0.291--0.695) & 100 & 7.05 & 17.9 & 31 / 31 / 31 & 0 / 0 / 0\\
\addlinespace[2pt]
P1-P & 0.5 & 32000 & 0.0 / 0.0 / 0.0 & 0.11 (0.084--0.181) & 115 & 1.97 & 6.46 & 31 / 31 / 31 & 0 / 0 / 0\\
\addlinespace[2pt]
P1-P & 0.3 & 2000 & 0.0 / 1.6 / 74.0 & 17 (7.41--21.9) & 13.6 & 45.2 & 146 & 31 / 31 / 10 & 5 / 27 / 31\\
\addlinespace[2pt]
P1-P & 0.3 & 8000 & 0.0 / 0.0 / 0.0 & 1.53 (0.624--2.05) & 75.3 & 16.5 & 46.7 & 31 / 31 / 31 & 0 / 0 / 0\\
\addlinespace[2pt]
P1-P & 0.3 & 32000 & 0.0 / 0.0 / 0.0 & 0.182 (0.116--0.272) & 105 & 5.27 & 12.7 & 31 / 31 / 31 & 0 / 0 / 0\\
\bottomrule\end{tabular}}\end{adjustbox}\end{table}

{\scriptsize\setlength{\tabcolsep}{3pt}
\begin{longtable}{lrrrrlrrrrrrrrrl}
\caption{The named configurations at every $(C,n)$. \emph{disp.}: intrinsic signed along-fiber bias; \emph{spread}: replicate SD of the intrinsic position; both in \% of the reference arc. $R$: replicates (pooled runs at $n=2000$). \emph{dist/$|e|$}: median distance to the fiber over the distance $|e|=|\hat\theta-\theta_0|$ to the truth. \emph{$\le0.02$}: share of replicates within $0.02$ of the fiber in $\theta$. Flags: \textsc{bs} (sign opposite to the P1-cluster branch pattern), \textsc{bd} (breakdown, mean degree 45), n.s.\ (not significant at 2 SE), boundary (configuration at the arc end).}\label{tab:keycells}\\
\toprule est. & $C$ & $n$ & deg. & $R$ & config & $w_{\min}$ & disp. & $\pm$SE & spread & RMSE & $\Delta m_3$ & $\hat w/w$ & dist/$|e|$ & $\le0.02$ & flags\\\midrule\endfirsthead
\toprule est. & $C$ & $n$ & deg. & $R$ & config & $w_{\min}$ & disp. & $\pm$SE & spread & RMSE & $\Delta m_3$ & $\hat w/w$ & dist/$|e|$ & $\le0.02$ & flags\\\midrule\endhead\bottomrule\endfoot
P1-c & 0.75 & 2000 & 281 & 400 & LO & 0.2111 & +8.62 & 0.069 & 1.37 & 8.73 & +3.78 & 1.373 & 0.3486 & 0 & --\\
P1-c & 0.75 & 2000 & 281 & 400 & pair A & 0.2552 & +9.23 & 0.073 & 1.46 & 9.34 & +3.48 & 1.342 & 0.3659 & 0 & --\\
P1-c & 0.75 & 2000 & 281 & 400 & MID & 0.314 & -8.26 & 0.092 & 1.85 & 8.47 & -2.44 & 1.305 & 0.5534 & 0 & --\\
P1-c & 0.75 & 2000 & 281 & 400 & HI & 0.03754 & -48.8 & 0.3 & 5.92 & 49.2 & -21.1 & 9.07 & 0.09744 & 2 & --\\
P1-c & 0.75 & 2000 & 281 & 400 & pair B & 0.03613 & -50.3 & 0.29 & 5.75 & 50.6 & -21.6 & 9.536 & 0.0902 & 2.25 & --\\
\addlinespace[2pt]
P1-c & 0.75 & 8000 & 1126 & 200 & LO & 0.2111 & +1.09 & 0.027 & 0.38 & 1.16 & +0.533 & 1.038 & 0.4616 & 100 & --\\
P1-c & 0.75 & 8000 & 1126 & 200 & pair A & 0.2552 & +1.42 & 0.032 & 0.456 & 1.49 & +0.586 & 1.048 & 0.3778 & 100 & --\\
P1-c & 0.75 & 8000 & 1125 & 200 & MID & 0.314 & -2.11 & 0.036 & 0.51 & 2.17 & -0.655 & 1.077 & 0.4751 & 100 & --\\
P1-c & 0.75 & 8000 & 1127 & 200 & HI & 0.03754 & -7.49 & 0.1 & 1.45 & 7.63 & -3.93 & 1.405 & 0.06409 & 100 & --\\
P1-c & 0.75 & 8000 & 1126 & 200 & pair B & 0.03613 & -7.74 & 0.11 & 1.49 & 7.88 & -4.04 & 1.41 & 0.06059 & 100 & --\\
\addlinespace[2pt]
P1-c & 0.75 & 32000 & 4500 & 200 & LO & 0.2111 & +0.111 & 0.012 & 0.167 & 0.201 & +0.0549 & 1.001 & 0.8005 & 100 & --\\
P1-c & 0.75 & 32000 & 4500 & 200 & pair A & 0.2552 & +0.107 & 0.016 & 0.221 & 0.246 & +0.0451 & 1.001 & 0.814 & 100 & --\\
P1-c & 0.75 & 32000 & 4499 & 200 & MID & 0.314 & -0.261 & 0.016 & 0.224 & 0.344 & -0.0827 & 1.009 & 0.5411 & 100 & --\\
P1-c & 0.75 & 32000 & 4502 & 200 & HI & 0.03754 & -0.238 & 0.014 & 0.192 & 0.306 & -0.125 & 1.014 & 0.5489 & 100 & --\\
P1-c & 0.75 & 32000 & 4500 & 200 & pair B & 0.03613 & -0.26 & 0.013 & 0.188 & 0.321 & -0.136 & 1.008 & 0.4755 & 100 & --\\
\addlinespace[2pt]
P1-c & 0.5 & 2000 & 125 & 400 & LO & 0.2111 & +28.2 & 0.1 & 2.06 & 28.3 & +10.3 & 2.242 & 0.3649 & 0 & --\\
P1-c & 0.5 & 2000 & 125 & 400 & pair A & 0.2552 & +25.3 & 0.1 & 2.04 & 25.4 & +8.51 & 1.9 & 0.4082 & 0 & --\\
P1-c & 0.5 & 2000 & 125 & 400 & MID & 0.314 & -10 & 0.11 & 2.15 & 10.3 & -2.93 & 1.402 & 0.7795 & 0 & --\\
P1-c & 0.5 & 2000 & 125 & 400 & HI & 0.03754 & -55.4 & 0.16 & 3.15 & 55.4 & -22.6 & 11.14 & 0.2545 & 0 & --\\
P1-c & 0.5 & 2000 & 125 & 400 & pair B & 0.03613 & -56.7 & 0.15 & 3.09 & 56.8 & -23.1 & 11.6 & 0.2506 & 0 & --\\
\addlinespace[2pt]
P1-c & 0.5 & 8000 & 500 & 200 & LO & 0.2111 & +4.29 & 0.034 & 0.477 & 4.32 & +2 & 1.182 & 0.3301 & 99.5 & --\\
P1-c & 0.5 & 8000 & 500 & 200 & pair A & 0.2552 & +5.09 & 0.041 & 0.583 & 5.13 & +2.01 & 1.189 & 0.3319 & 76.5 & --\\
P1-c & 0.5 & 8000 & 500 & 200 & MID & 0.314 & -6.69 & 0.054 & 0.767 & 6.73 & -2 & 1.237 & 0.4145 & 0 & --\\
P1-c & 0.5 & 8000 & 500 & 200 & HI & 0.03754 & -44.6 & 0.27 & 3.79 & 44.7 & -20 & 7.789 & 0.03353 & 86.5 & --\\
P1-c & 0.5 & 8000 & 500 & 200 & pair B & 0.03613 & -46 & 0.27 & 3.75 & 46.1 & -20.5 & 8.026 & 0.03229 & 83.5 & --\\
\addlinespace[2pt]
P1-c & 0.5 & 32000 & 2002 & 200 & LO & 0.2111 & +0.458 & 0.014 & 0.194 & 0.497 & +0.226 & 1.012 & 0.6013 & 100 & --\\
P1-c & 0.5 & 32000 & 2001 & 200 & pair A & 0.2552 & +0.636 & 0.015 & 0.211 & 0.67 & +0.266 & 1.019 & 0.4467 & 100 & --\\
P1-c & 0.5 & 32000 & 2001 & 200 & MID & 0.314 & -1.11 & 0.017 & 0.239 & 1.13 & -0.348 & 1.041 & 0.4362 & 100 & --\\
P1-c & 0.5 & 32000 & 1999 & 200 & HI & 0.03754 & -2.57 & 0.029 & 0.415 & 2.6 & -1.35 & 1.112 & 0.1074 & 100 & --\\
P1-c & 0.5 & 32000 & 2001 & 200 & pair B & 0.03613 & -2.62 & 0.032 & 0.454 & 2.66 & -1.36 & 1.11 & 0.106 & 100 & --\\
\addlinespace[2pt]
P1-c & 0.3 & 2000 & 45 & 400 & LO & 0.2111 & +69.3 & 0.84 & 16.9 & 71.3 & +27.5 & 1.794 & 0.5458 & 0 & \textsc{bd}\\
P1-c & 0.3 & 2000 & 45 & 400 & pair A & 0.2552 & +64.8 & 0.88 & 17.5 & 67.2 & +24.9 & 1.478 & 0.5817 & 0 & \textsc{bd}\\
P1-c & 0.3 & 2000 & 45 & 400 & MID & 0.314 & +21.5 & 0.85 & 16.9 & 27.3 & +9.26 & 1.2 & 0.9789 & 0 & \textsc{bs}\\
P1-c & 0.3 & 2000 & 45 & 400 & HI & 0.03754 & -24.8 & 0.87 & 17.4 & 30.3 & -12.7 & 9.973 & 0.9459 & 0 & --\\
P1-c & 0.3 & 2000 & 45 & 400 & pair B & 0.03613 & -26.3 & 0.86 & 17.1 & 31.4 & -13.4 & 10.4 & 0.9324 & 0 & \textsc{bd}\\
\addlinespace[2pt]
P1-c & 0.3 & 8000 & 180 & 200 & LO & 0.2111 & +21.6 & 0.075 & 1.07 & 21.7 & +8.3 & 1.963 & 0.3162 & 0 & --\\
P1-c & 0.3 & 8000 & 180 & 200 & pair A & 0.2552 & +19.8 & 0.07 & 0.99 & 19.8 & +6.85 & 1.739 & 0.3583 & 0 & --\\
P1-c & 0.3 & 8000 & 180 & 200 & MID & 0.314 & -11.5 & 0.074 & 1.04 & 11.5 & -3.33 & 1.422 & 0.6074 & 0 & --\\
P1-c & 0.3 & 8000 & 180 & 200 & HI & 0.03754 & -57 & 0.11 & 1.6 & 57 & -22.9 & 11.26 & 0.1627 & 0 & --\\
P1-c & 0.3 & 8000 & 180 & 200 & pair B & 0.03613 & -58.3 & 0.11 & 1.49 & 58.3 & -23.4 & 11.78 & 0.157 & 0 & --\\
\addlinespace[2pt]
P1-c & 0.3 & 32000 & 720 & 200 & LO & 0.2111 & +2.65 & 0.016 & 0.232 & 2.66 & +1.26 & 1.107 & 0.3294 & 100 & --\\
P1-c & 0.3 & 32000 & 720 & 200 & pair A & 0.2552 & +3.22 & 0.017 & 0.245 & 3.23 & +1.3 & 1.116 & 0.3226 & 100 & --\\
P1-c & 0.3 & 32000 & 720 & 200 & MID & 0.314 & -4.96 & 0.024 & 0.341 & 4.98 & -1.51 & 1.176 & 0.3808 & 13.5 & --\\
P1-c & 0.3 & 32000 & 720 & 200 & HI & 0.03754 & -31.4 & 0.33 & 4.67 & 31.7 & -15.6 & 4.429 & 0.03061 & 100 & --\\
P1-c & 0.3 & 32000 & 720 & 200 & pair B & 0.03613 & -33.1 & 0.31 & 4.42 & 33.4 & -16.3 & 4.666 & 0.02988 & 100 & --\\
\midrule
P1-P & 0.75 & 2000 & 281 & 400 & LO & 0.2111 & +74.4 & 0.54 & 10.8 & 75.1 & +31 & 0.9706 & 0.182 & 0 & --\\
P1-P & 0.75 & 2000 & 281 & 400 & pair A & 0.2552 & +71.1 & 0.55 & 10.9 & 71.9 & +29.3 & 0.7744 & 0.1906 & 0 & --\\
P1-P & 0.75 & 2000 & 281 & 400 & MID & 0.314 & +35.6 & 0.42 & 8.43 & 36.6 & +18.1 & 0.5049 & 0.3252 & 0 & \textsc{bs}\\
P1-P & 0.75 & 2000 & 281 & 400 & HI & 0.03754 & -1.64 & 0.22 & 4.32 & 4.63 & -0.862 & 2.776 & 0.8827 & 0 & --\\
P1-P & 0.75 & 2000 & 281 & 400 & pair B & 0.03613 & -2.62 & 0.21 & 4.13 & 4.89 & -1.36 & 2.935 & 0.9304 & 0 & --\\
\addlinespace[2pt]
P1-P & 0.75 & 8000 & 1126 & 200 & LO & 0.2111 & +25.2 & 0.13 & 1.83 & 25.3 & +9.42 & 2.166 & 0.1195 & 0 & --\\
P1-P & 0.75 & 8000 & 1126 & 200 & pair A & 0.2552 & +24.3 & 0.12 & 1.68 & 24.4 & +8.22 & 1.911 & 0.124 & 0 & --\\
P1-P & 0.75 & 8000 & 1125 & 200 & MID & 0.314 & +6.16 & 0.089 & 1.26 & 6.28 & +2.1 & 0.8616 & 0.5281 & 0 & \textsc{bs}\\
P1-P & 0.75 & 8000 & 1127 & 200 & HI & 0.03754 & -7.79 & 0.092 & 1.31 & 7.9 & -4.08 & 1.947 & 0.2778 & 15 & --\\
P1-P & 0.75 & 8000 & 1126 & 200 & pair B & 0.03613 & -8.25 & 0.094 & 1.32 & 8.36 & -4.3 & 1.969 & 0.248 & 17.5 & --\\
\addlinespace[2pt]
P1-P & 0.75 & 32000 & 4500 & 200 & LO & 0.2111 & +6.07 & 0.017 & 0.235 & 6.07 & +2.75 & 1.277 & 0.1432 & 100 & --\\
P1-P & 0.75 & 32000 & 4500 & 200 & pair A & 0.2552 & +6.13 & 0.019 & 0.272 & 6.13 & +2.39 & 1.237 & 0.1328 & 100 & --\\
P1-P & 0.75 & 32000 & 4499 & 200 & MID & 0.314 & +1.06 & 0.015 & 0.216 & 1.08 & +0.339 & 0.9759 & 0.6626 & 100 & \textsc{bs}\\
P1-P & 0.75 & 32000 & 4502 & 200 & HI & 0.03754 & -3.2 & 0.019 & 0.262 & 3.21 & -1.68 & 1.256 & 0.1442 & 100 & --\\
P1-P & 0.75 & 32000 & 4500 & 200 & pair B & 0.03613 & -3.36 & 0.018 & 0.254 & 3.37 & -1.75 & 1.253 & 0.1268 & 100 & --\\
\addlinespace[2pt]
P1-P & 0.5 & 2000 & 125 & 400 & LO & 0.2111 & +113 & 0.36 & 7.25 & 114 & +60.9 & 0.5349 & 0.1193 & 0 & --\\
P1-P & 0.5 & 2000 & 125 & 400 & pair A & 0.2552 & +109 & 0.35 & 7.06 & 109 & +57.5 & 0.4365 & 0.1258 & 0 & --\\
P1-P & 0.5 & 2000 & 125 & 400 & MID & 0.314 & +66.6 & 0.32 & 6.39 & 66.9 & +38.6 & 0.3374 & 0.1875 & 0 & \textsc{bs}\\
P1-P & 0.5 & 2000 & 125 & 400 & HI & 0.03754 & +20.4 & 0.23 & 4.66 & 20.9 & +10.6 & 2.571 & 0.4374 & 0 & \textsc{bs}\\
P1-P & 0.5 & 2000 & 125 & 400 & pair B & 0.03613 & +19.5 & 0.23 & 4.51 & 20 & +10.1 & 2.641 & 0.444 & 0 & \textsc{bs}\\
\addlinespace[2pt]
P1-P & 0.5 & 8000 & 500 & 200 & LO & 0.2111 & +50.1 & 0.26 & 3.7 & 50.2 & +17.6 & 1.521 & 0.1562 & 0 & --\\
P1-P & 0.5 & 8000 & 500 & 200 & pair A & 0.2552 & +47.5 & 0.26 & 3.69 & 47.6 & +15.9 & 1.19 & 0.1689 & 0 & --\\
P1-P & 0.5 & 8000 & 500 & 200 & MID & 0.314 & +18 & 0.24 & 3.38 & 18.3 & +7.35 & 0.6658 & 0.4625 & 0 & \textsc{bs}\\
P1-P & 0.5 & 8000 & 500 & 200 & HI & 0.03754 & -7.83 & 0.19 & 2.67 & 8.27 & -4.11 & 2.726 & 0.5492 & 0 & --\\
P1-P & 0.5 & 8000 & 500 & 200 & pair B & 0.03613 & -8.44 & 0.18 & 2.59 & 8.82 & -4.4 & 2.796 & 0.5169 & 0 & --\\
\addlinespace[2pt]
P1-P & 0.5 & 32000 & 2002 & 200 & LO & 0.2111 & +14.9 & 0.03 & 0.425 & 15 & +6.1 & 1.687 & 0.1261 & 56.5 & --\\
P1-P & 0.5 & 32000 & 2001 & 200 & pair A & 0.2552 & +14.9 & 0.037 & 0.518 & 14.9 & +5.33 & 1.575 & 0.1241 & 80.5 & --\\
P1-P & 0.5 & 32000 & 2001 & 200 & MID & 0.314 & +3 & 0.027 & 0.389 & 3.02 & +0.986 & 0.9319 & 0.6043 & 0 & \textsc{bs}\\
P1-P & 0.5 & 32000 & 1999 & 200 & HI & 0.03754 & -6.36 & 0.033 & 0.471 & 6.38 & -3.33 & 1.616 & 0.1924 & 100 & --\\
P1-P & 0.5 & 32000 & 2001 & 200 & pair B & 0.03613 & -6.6 & 0.035 & 0.489 & 6.62 & -3.44 & 1.623 & 0.1785 & 100 & --\\
\addlinespace[2pt]
P1-P & 0.3 & 2000 & 45 & 400 & LO & 0.2111 & +119 & 0 & 0 & 119 & +65.3 & 0.3441 & 0.4986 & 0 & \textsc{bd}\\
P1-P & 0.3 & 2000 & 45 & 400 & pair A & 0.2552 & +114 & 0 & 0 & 114 & +61.7 & 0.2741 & 0.5121 & 0 & \textsc{bd}\\
P1-P & 0.3 & 2000 & 45 & 400 & MID & 0.314 & +71.1 & 5.3e-16 & 1.05e-14 & 71.1 & +41.5 & 0.2329 & 0.5734 & 0 & \textsc{bs}\\
P1-P & 0.3 & 2000 & 45 & 400 & HI & 0.03754 & +23.1 & 3.9e-16 & 7.89e-15 & 23.1 & +12 & 1.93 & 0.7848 & 0 & \textsc{bs}\\
P1-P & 0.3 & 2000 & 45 & 400 & pair B & 0.03613 & +22.1 & 1.3e-16 & 2.63e-15 & 22.1 & +11.4 & 1.952 & 0.7971 & 0 & \textsc{bs} \textsc{bd}\\
\addlinespace[2pt]
P1-P & 0.3 & 8000 & 180 & 200 & LO & 0.2111 & +97.9 & 0.41 & 5.84 & 98.1 & +49 & 0.6818 & 0.1398 & 0 & --\\
P1-P & 0.3 & 8000 & 180 & 200 & pair A & 0.2552 & +93.9 & 0.41 & 5.83 & 94.1 & +46.4 & 0.5499 & 0.1422 & 0 & --\\
P1-P & 0.3 & 8000 & 180 & 200 & MID & 0.314 & +54 & 0.44 & 6.27 & 54.4 & +30.3 & 0.4053 & 0.2115 & 0 & \textsc{bs}\\
P1-P & 0.3 & 8000 & 180 & 200 & HI & 0.03754 & +12.6 & 0.38 & 5.41 & 13.7 & +6.58 & 2.839 & 0.567 & 0 & \textsc{bs}\\
P1-P & 0.3 & 8000 & 180 & 200 & pair B & 0.03613 & +11.2 & 0.35 & 4.95 & 12.2 & +5.8 & 2.939 & 0.5994 & 0 & \textsc{bs}\\
\addlinespace[2pt]
P1-P & 0.3 & 32000 & 720 & 200 & LO & 0.2111 & +38.9 & 0.096 & 1.35 & 38.9 & +13.7 & 1.965 & 0.1348 & 0 & --\\
P1-P & 0.3 & 32000 & 720 & 200 & pair A & 0.2552 & +36.8 & 0.1 & 1.42 & 36.8 & +12.1 & 1.523 & 0.1444 & 0 & --\\
P1-P & 0.3 & 32000 & 720 & 200 & MID & 0.314 & +10.9 & 0.071 & 1 & 11 & +3.99 & 0.7692 & 0.5152 & 0 & \textsc{bs}\\
P1-P & 0.3 & 32000 & 720 & 200 & HI & 0.03754 & -9.6 & 0.076 & 1.08 & 9.67 & -5.03 & 2.504 & 0.3894 & 0 & --\\
P1-P & 0.3 & 32000 & 720 & 200 & pair B & 0.03613 & -10 & 0.064 & 0.898 & 10 & -5.21 & 2.539 & 0.3647 & 0 & --\\
\end{longtable}}

\paragraph{Provenance.}

Eight runs on $64$ CPUs and $6\times$ RTX 6000 Ada: about $10$ wall-hours for the first seven and a few more for the eighth, whose Bayesian chains were capped at $3$ hours.
Master seed $20260806$; per-replicate seeds $\mathrm{sha256}(\text{master}\,|\,
\text{tag})$ with disjoint tag prefixes across runs. Configurations loaded
bit-identically from the first run. The code passes $35$ unit tests, including
brute-force motif enumeration for $n\le12$; the suppression identity was
verified at $50$ digits in three independent ways; and every cell has positive
controls.
Triangle and friendship-graph accumulation use integer arithmetic (the
separator signal is below float32 epsilon). Python 3.12.3, numpy 1.26.4, scipy
1.11.4, pandas 2.1.4, mpmath 1.3.0, torch 2.7.0, CUDA 12.8.

\end{document}